\documentclass[11pt]{article}

\usepackage[final]{acl}

\usepackage{times}
\usepackage{latexsym}

\usepackage[T1]{fontenc}

\usepackage[utf8]{inputenc}

\usepackage{microtype}

\usepackage{inconsolata}

\usepackage{graphicx}

\usepackage{adjustbox}
\usepackage[table]{xcolor}
\usepackage{booktabs, multirow, array, xcolor, caption, tabularx}
\usepackage{subcaption}
\usepackage{pgfplots}

\usepackage{tikz}
\usepackage[edges]{forest} 
\usepackage{arydshln} 

\usepackage{pgfplots}
\usepackage{makecell}

\usepackage{dsfont}

\usepackage{courier}
\usepackage[most]{tcolorbox}

\newcommand*{\img}[1]{%
    \raisebox{-.2\baselineskip}{%
        \includegraphics[
        height=\baselineskip,
        width=\baselineskip,
        keepaspectratio,
        ]{#1}%
    }%
}

\title{\img{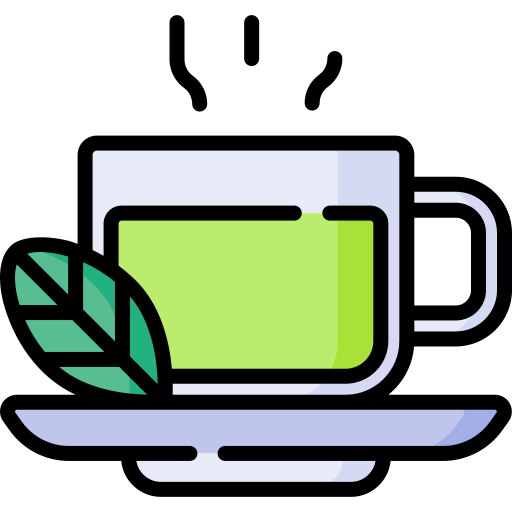} TEA-Bench: A Systematic Benchmarking of \\\underline{T}ool-enhanced \underline{E}motional Support Dialogue \underline{A}gent}

\author{Xingyu Sui, Yanyan Zhao\thanks{\ \ Corresponding author}, Yulin Hu, Jiahe Guo, Weixiang Zhao, Bing Qin\\
        Harbin Institute of Technology \\
        \texttt{\{xysui, wxzhao, yyzhao\}@ir.hit.edu.cn}}

\definecolor{blue1}{HTML}{0ebeff}
\definecolor{red1}{HTML}{ff42b3}

\definecolor{tkcolor}{RGB}{224,223,255}
\newtcolorbox{takeaways}[2][]{
	width=\columnwidth,
	colback = tkcolor, 
	colframe = tkcolor, 
	boxsep=0pt,left=10pt,right=10pt,top=2pt,bottom=3pt,
	fontupper=\linespread{0.9}\selectfont,
	title=#2,#1}

\newtcolorbox{promptbox}[2][]{
	width=\linewidth,
	colback = gray!8, 
	colframe = gray!8, 
	coltitle = black,
	colbacktitle = gray!8,            
	boxsep=0pt,left=10pt,right=10pt,top=0pt,bottom=0pt,
	fontupper=\linespread{0.9}\selectfont,
	fonttitle=\bfseries,
	title=#2,#1
}

\newtcolorbox{promptboxsm}[2][]{
	width=\linewidth,
	colback = gray!8, 
	colframe = gray!8, 
	coltitle = black,
	colbacktitle = gray!8,            
	boxsep=0pt,left=10pt,right=10pt,top=0pt,bottom=0pt,
	fontupper=\scriptsize\linespread{0.9}\selectfont,
	fonttitle=\bfseries,
	title=#2,#1
}

\newtcolorbox{examplebox}[1][]{
  width=\linewidth,
  colback = orange!6,
  colframe = orange!30,
  coltitle = black,
  boxrule = 0.4pt,
  boxsep=0pt,left=10pt,right=10pt,top=6pt,bottom=6pt,
  fontupper=\linespread{0.95}\selectfont,
  #1
}

\definecolor{blue2}{HTML}{398cd6}
\newcommand{\fstring}[1]{\textcolor{blue2}{\detokenize{{#1}}}}

\begin{document}
\maketitle

\begin{abstract}

Emotional Support Conversation requires not only affective expression but also grounded instrumental support to provide trustworthy guidance. However, existing ESC systems and benchmarks largely focus on affective support in text-only settings, overlooking how external tools can enable factual grounding and reduce hallucination in multi-turn emotional support. We introduce \textbf{\texttt{TEA-Bench}}, the first interactive benchmark for evaluating tool-augmented agents in ESC, featuring realistic emotional scenarios, an MCP-style tool environment, and process-level metrics that jointly assess the quality and factual grounding of emotional support.
Experiments on nine LLMs show that tool augmentation generally improves emotional support quality and reduces hallucination, but the gains are strongly capacity-dependent: stronger models use tools more selectively and effectively, while weaker models benefit only marginally. We further release \textbf{\texttt{TEA-Dialog}}, a dataset of tool-enhanced ESC dialogues, and find that supervised fine-tuning improves in-distribution support but generalizes poorly. Our results underscore the importance of tool use in building reliable emotional support agents. \footnote{\ \ Our code and data can be found in \url{https://github.com/XingYuSSS/TEA-Bench}.}

\end{abstract}
\section{Introduction}

\begin{figure}[t]
    \centering
    \includegraphics[width=\linewidth]{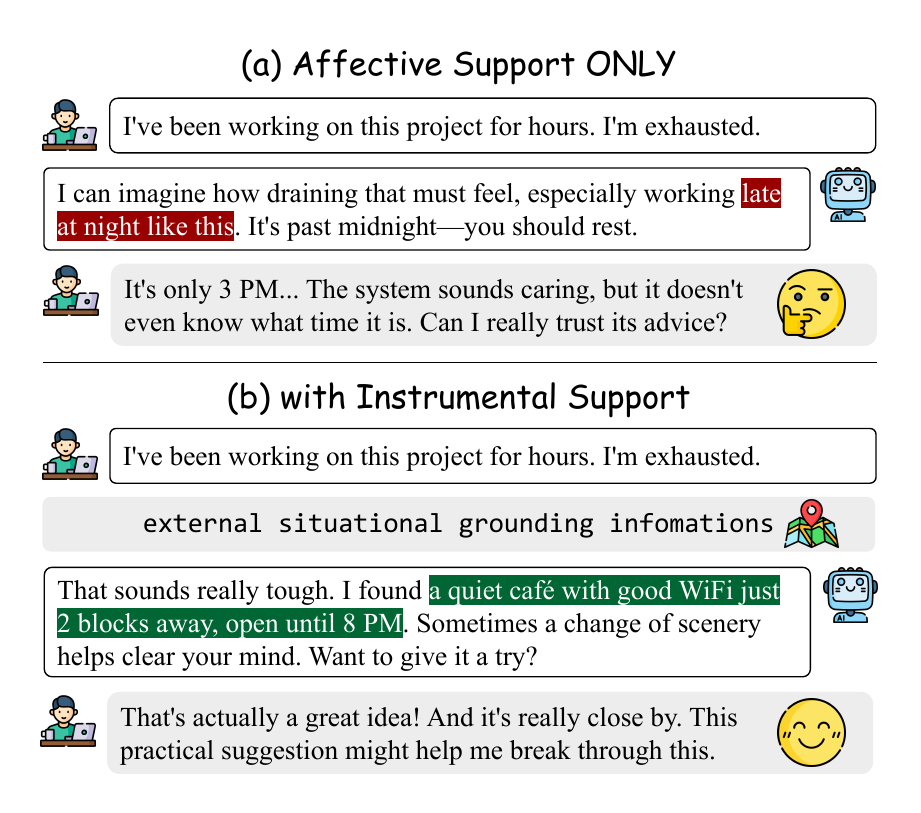}
    \caption{
Comparison of affective-only support vs. instrumental support. (a) Affective-only response \textcolor[HTML]{990000}{hallucinates details}, undermining trust. (b) Instrumental support response provides \textcolor[HTML]{006633}{verified, actionable suggestions}.
    }
    \label{fig:comp}
    \vspace{-0.5cm}
\end{figure}

In modern society, people increasingly experience emotional stress due to mounting pressures from work and daily life. Thus, the demand for Emotional Support Conversation (ESC) \citep{liu2021towards} has grown substantially, as they offer psychological relief, reassurance and guidance during moments of distress \citep{langford1997social, greene2003handbook, heaney2008social}.

Social support theory distinguishes two complementary types of support in ESC: \textbf{affective support}, which conveys empathy and care, and \textbf{instrumental support}, which offers concrete guidance and actionable assistance \citep{cutrona1990type, semmer2008emotional}. While affective support can be expressed through empathetic language alone, instrumental support requires accurate grounding in real-world situational information. Without such grounding, advice may become generic or factually incorrect, undermining user trust \citep{2017Social, shaikh2024grounding}.

However, existing ESC systems predominantly focus on affective support alone. Despite advances in linguistic fluency and empathetic expression, recent models operate largely in text-only settings with limited access to external contextual information \citep{zhao2023chatgpt, chen2023soulchat, farhat2024chatgpt}. As illustrated in Figure~\ref{fig:comp}(a), affective-only responses often contain unwarranted hallucinated contextual assumptions that damage credibility despite sounding caring. In contrast, Figure~\ref{fig:comp}(b) shows how responses grounded with instrumental support leverage verified external information to provide trustworthy, actionable suggestions that complement emotional validation.

Recent advances in tool-augmented large language models (LLMs), provide a promising pathway toward addressing this gap by allowing models to dynamically acquire external situational knowledge during interaction \citep{mialon2023augmentedlanguagemodelssurvey, qin2024tool}. Nevertheless, existing evaluation frameworks remain inadequate. ESC benchmarks predominantly assess text-only empathy \citep{zhao2024esc}, while evaluations of tool-augmented systems mainly focus on task-oriented performance \citep{zhang2024toolbehonest, li2023api, trivedi2024appworld}. As a result, the role of tools in enabling grounded, empathetic, and multi-turn emotional support remains poorly understood.

To address this gap, we introduce \texttt{TEA-Bench} (Tool-enhanced Emotional Support Dialogue Agent Benchmark), the first interactive benchmark for evaluating tool-augmented agents in ESC. \texttt{TEA-Bench} investigates how access to external tools facilitates grounded, empathetic, and context-aware support in multi-turn interactions. The benchmark comprises fine-grained emotional support scenarios adapted from ExTES \citep{zheng2024self}, a realistic MCP-based tool environment, and a simulated user that provides structured feedback, including reactions to hallucinated or inappropriate responses. Crucially, \texttt{TEA-Bench} enables process-level analysis through complementary metrics that assess both ESC quality and factual grounding, capturing model behavior across conversational turns.

Using \texttt{TEA-Bench}, we evaluate nine contemporary LLMs, including four closed-source and five open-source models. We identify capacity-dependent patterns in tool-enhanced emotional support: tool augmentation improves ESC performance and reduces hallucination, but the gains depend on models’ ability to invoke and integrate tools effectively. Stronger models leverage precise, selective tool usage, while mid-capability models rely on more frequent calls to achieve comparable grounding; weaker models struggle to benefit and show limited empathy gains. Further analysis reveals a positive relationship between tool usage and hallucination mitigation, with substantial efficiency differences across model scales.

We also curate \texttt{TEA-Dialog}, a high-quality dataset of grounded, tool-enhanced dialogues derived from \texttt{TEA-Bench} interactions. While supervised fine-tuning on this dataset improves empathetic performance in familiar scenarios, it exhibits limited generalization and may increase hallucination under distribution shift, highlighting both the promise and challenges of training reliable tool-augmented emotional support agents.

This work makes three main contributions: (1) we articulate the need for grounded instrumental support in emotional support conversations and highlight the lack of evaluation frameworks that capture how external tools enable context-aware, trustworthy empathy in multi-turn ESC; (2) we introduce \texttt{TEA-Bench}, the first interactive benchmark designed to evaluate tool-augmented emotional support agents, featuring realistic scenarios, an MCP-style tool environment, and process-level metrics for empathy and hallucination; and (3) through extensive experiments on nine contemporary LLMs, we uncover capacity-dependent patterns in tool utilization and hallucination mitigation, and release \texttt{TEA-Dialog}, a high-quality dataset of grounded, tool-enhanced ESC dialogues to support future research and training.

\section{Related Works}

\paragraph{Emotional Support Conversation.}
Emotional Support Conversations (ESC) \citep{liu2021towards} study interactions between a seeker experiencing emotional distress and a supporter aiming to alleviate emotional intensity through appropriate conversational strategies. Early work explored task-specific architectures and modeling techniques, such as hierarchical graph networks \citep{peng2022control, zhao2022cauain}, commonsense knowledge integration \citep{tu2022misc}, and joint emotion–semantic modeling \citep{ zhao2023transesc}. With the emergence of large language models (LLMs), recent approaches leverage pretrained models via supervised fine-tuning on curated ESC datasets, improving multi-turn coherence and empathetic quality without architectural modification \citep{liu2023chatcounselor, chen2023soulchat, qiu2023smile}. Beyond supervised learning, \citet{zhao2025chain} propose a Chain of Strategy Optimization (CSO) framework that uses Monte Carlo Tree Search to generate high-quality preference data and applies preference optimization, further enhancing ESC performance. However, these approaches primarily operate in text-only settings and do not consider the role of external grounding or tool use in emotional support.

\paragraph{Benchmarks for Tool-augmented LLM.}
Tool-augmented large language models enable interaction with external tools and real-world information sources, substantially extending their capabilities \cite{mialon2023augmentedlanguagemodelssurvey, qin2024tool}. Prior benchmarks mainly evaluate task-oriented tool use, focusing on accurate invocation, execution correctness, and task completion \cite{zhang2024toolbehonest, li2023api, trivedi2024appworld, liu2025agentbenchevaluatingllmsagents, han2025nestools}. However, these evaluations are largely designed for instruction-following or problem-solving scenarios and pay limited attention to human-centered interaction. How tool grounding should adapt to users’ emotional states and situational constraints in multi-turn supportive dialogue remains underexplored, motivating dedicated evaluation frameworks for emotional support settings.

\section{TEA-Bench}
\label{sec:teabench}

\begin{figure}[t]
    \centering
    \includegraphics[width=\linewidth]{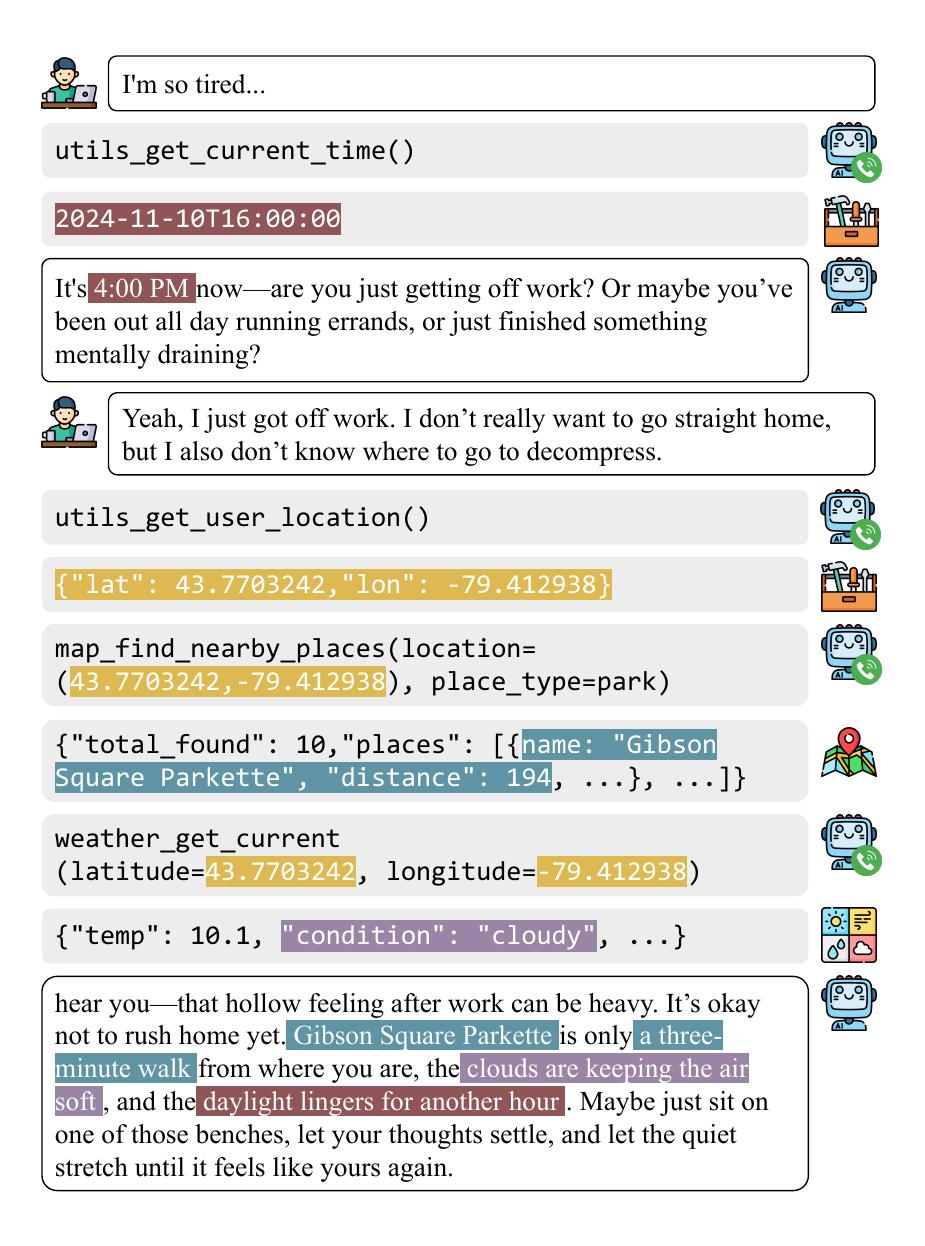}
    \caption{
Example illustrating a tool-augmented ESC under our benchmark setting. The assistant dynamically invokes external tools to gather contextual information, enabling it to offer emotionally resonant and actionable support rather than generic reassurance.
    }
    \label{fig:case}
    \vspace{-0.5cm}
\end{figure}

\begin{figure*}[t]
    \centering
    \includegraphics[width=2\columnwidth]{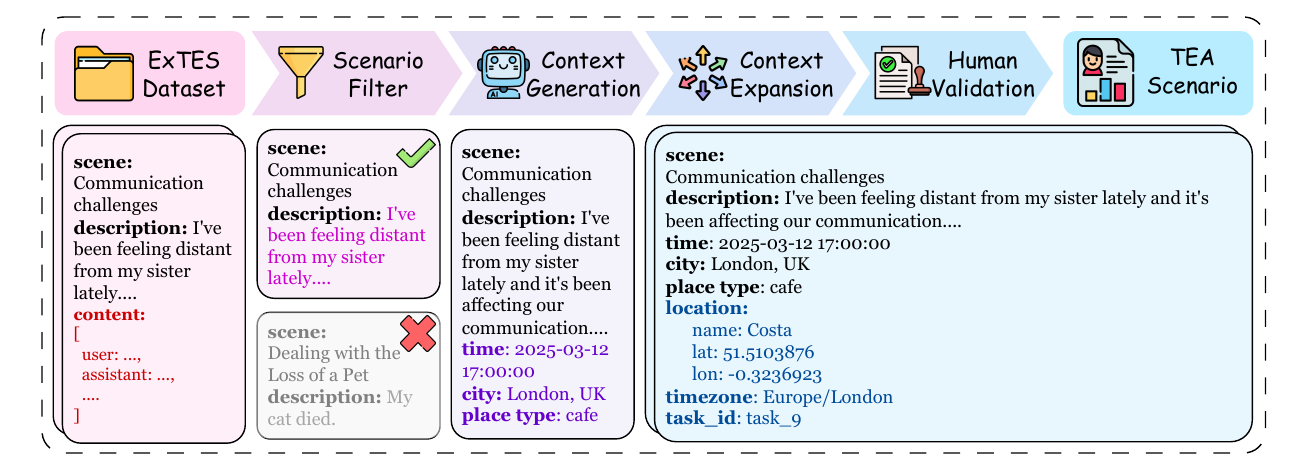}
    \caption{
       Overview of \texttt{TEA-Scenario} construction pipeline. We filter emotionally rich scenarios, generate latent spatiotemporal attributes via LLM, ground them through map-based APIs, and validate through human review.
    }
    \label{fig:data}
    \vspace{-0.5cm}
\end{figure*}

\subsection{Task Definition}
\label{sec:task_definition} 

Figure~\ref{fig:case} provides an illustrative example of a tool-augmented emotional support conversation under our benchmark. \texttt{TEA-Bench} evaluates agents in emotional support conversations between a user and an agent. At each turn, the agent first decides whether external information is needed and may optionally invoke tools to retrieve situational or factual information. Any retrieved tool outputs are incorporated into the agent’s context. Based on the dialogue history and this contextual information, the agent then generates an appropriate response. Tool use serves as auxiliary support for grounding rather than completing an explicit task.

\paragraph{Asymmetric Information Access.}
The agent observes the full interaction history including all tool invocations and outputs. In contrast, users and evaluators observe only the natural language utterances, mirroring real-world scenarios where internal information retrieval is hidden from users.

\paragraph{Evaluation Objectives.}
Agents are evaluated on: (1) \textbf{empathetic support quality}, consistently providing emotionally appropriate and helpful responses; and (2) \textbf{factual grounding}, ensuring concrete claims are traceable to user-provided information or tool observations, avoiding hallucinated content. Unlike task-oriented dialogues, no explicit terminal goal is defined, reflecting the open-ended nature of emotional support.

\subsection{Scenario Construction}
\label{sec:scenario_construction}

We construct 81 grounded emotional support scenarios, referred to as \texttt{TEA-Scenarios}, from the ExTES dataset \citep{zheng2024self} through a four-stage pipeline shown in Figure~\ref{fig:data}. Our goal is to enrich each scenario with retrievable spatiotemporal context, enabling agents to provide factually grounded support while preserving the authenticity of emotional support interactions.

\paragraph{Scenario Filtering.}
We retain ExTES scenarios with situation descriptions exceeding 30 words. This threshold ensures sufficient contextual richness for both empathetic reasoning and meaningful tool usage, filtering out underspecified cases where grounding opportunities would be limited.

\paragraph{Latent Context Generation.}
For each retained scenario, we use an LLM to infer implicit situational attributes that are relevant but not explicitly stated in emotional support contexts, including the user’s local time, approximate city-level location, and the type of place the user is in, such as a workplace or public environment. These attributes are generated by conditioning on the original scenario description and serve as latent contextual variables that agents can retrieve through tools. The generation prompt is provided in Appendix~\ref{app:latent_context_prompt}.

\paragraph{Context Grounding.}
The inferred location information is subsequently grounded using map-based API queries to obtain concrete geographic data, including precise coordinates, verified place names, and corresponding time zones. Each scenario is assigned a unique identifier to ensure consistent and reliable information retrieval.

\paragraph{Human Validation.}
All constructed scenarios undergo manual review by the authors to ensure quality and realism. We filter out cases with implausible location-situation pairings or internally inconsistent temporal information. Only scenarios that are coherent, contextually realistic, emotionally plausible, and suitable for emotional support interactions are retained in the final benchmark.

\begin{figure*}[t]
    \centering
    \includegraphics[width=2\columnwidth]{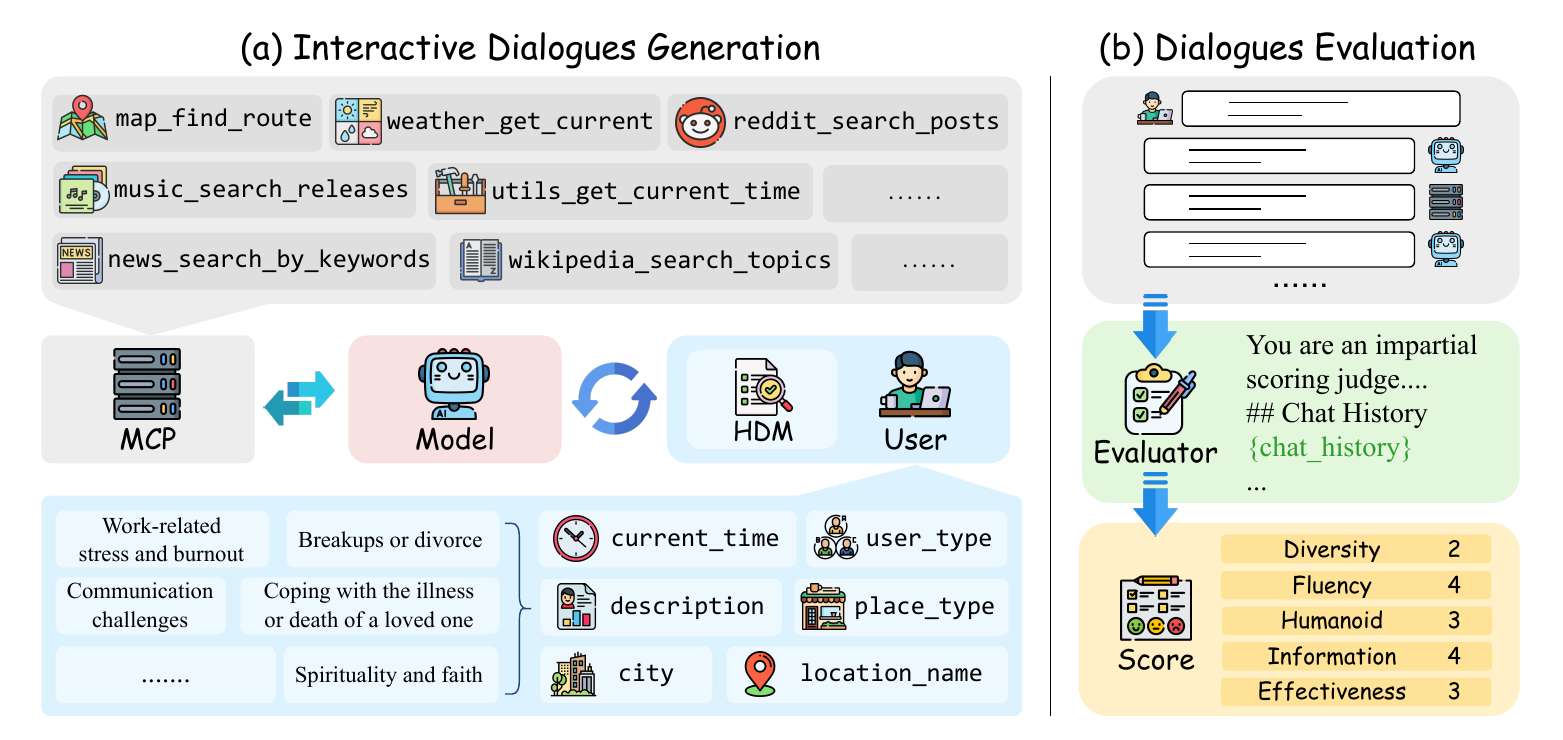}
    \caption{
Overview of \texttt{TEA-Bench}. The evaluated agent engages in multi-turn emotional support dialogues with a simulated user and may invoke external tools via the Model Context Protocol (MCP). A Hallucination Detection Module verifies factual grounding in agent responses based on dialogue history and tool observations. Complete dialogues are evaluated using the TEA score and factuality metrics.
    }
    \label{fig:main}
    \vspace{-0.5cm}
\end{figure*}

\subsection{Tool Environment}
\label{sec:tool_environment}

To support grounded instrumental support, \texttt{TEA-Bench} provides a diverse tool environment that allows agents to retrieve contextual information beyond the dialogue history.

\paragraph{Tool Design.}
The environment includes 31 tools across seven categories: \texttt{Reddit}, \texttt{Map}, \texttt{Utils}, \texttt{Weather}, \texttt{News}, \texttt{Wikipedia}, and \texttt{Music}. Appendix~\ref{app:tool_sources} details the complete tool list and their data sources. Each category serves a distinct grounding function. \texttt{Reddit} tools retrieve shared experiences from online communities. \texttt{Map} and \texttt{Weather} tools provide spatial and environmental information. \texttt{News} tools offer situational awareness. \texttt{Wikipedia} tools supply encyclopedic knowledge. \texttt{Music} tools support affective content recommendation. All tools expose real-world data sources via Model Context Protocol (MCP).

\paragraph{Scenario-Aware Execution.}
Time-sensitive tools (\texttt{Reddit}, \texttt{Weather}, \texttt{News}) execute in a scenario-aware manner to ensure reproducibility while preserving temporal realism. Tool calls implicitly condition on the timestamp of the current \texttt{TEA-Scenario} rather than the system clock. They return results corresponding to that specific moment. This mechanism provides a consistent external environment across agents while maintaining realistic time-dependent dynamics.

\paragraph{Context Retrieval Utilities.}
The \texttt{Utils} category retrieves scenario-specific contextual information without calling external APIs. These utilities directly access the latent attributes from Section~\ref{sec:scenario_construction}, including the user's local time, geographic coordinates, and current place. This simulates device-level information access that agents would have in real deployment. Agents can obtain grounded context through \texttt{Utils} before invoking downstream tools like \texttt{Map} or \texttt{Weather}.

\paragraph{Open-Ended Tool Usage.}
Tool use is entirely optional and unsupervised. Agents may invoke any subset of tools at any turn. They must infer from conversational context whether, when, and which tools to use. This reflects realistic deployment settings where appropriate instrumental support must be judged by the agent.

\subsection{Interactive Evaluation Framework}
\label{sec:interactive_framework}

\texttt{TEA-Bench} evaluates agents through interactive multi-turn dialogues with a simulated user. As shown in Figure~\ref{fig:main}(a), agents may invoke tools at any turn to retrieve contextual information. The agent is follows a fixed system prompt that encourages proactive tool usage and concise responses. Appendix~\ref{app:sys_prompt} provides the full prompt.

\paragraph{Tool Interaction.}
Tool usage is entirely controlled by the agent's native function-calling capability. All MCP tools are exposed as callable functions without prompt-level supervision. At each turn, the agent may generate a response or invoke tools. Tool observations are appended to the agent's context. The agent may continue calling tools or produce a final response.

\paragraph{Hallucination Detection Module (HDM).}
We introduce a Hallucination Detection Module (HDM) to identify whether factual content in agent responses is grounded in available context. A response is grounded if all factual entities (including locations, times, events, external conditions) can be traced to user-provided information or tool observations in the dialogue history.

Two types of content are excluded from HDM detection. First, commonsense elaborations that do not introduce new factual entities. Second, general emotional support or lifestyle suggestions that do not rely on external world states. The HDM employs a prompted language model that verifies entity provenance against dialogue and tool context. Appendix~\ref{app:hdm_prompt} provides the detection prompt.

\paragraph{User Simulation.}
User behavior is simulated using a language model conditioned on scenario attributes from Section~\ref{sec:scenario_construction}, including local time, geographic context, user type, and the emotional situation description. The simulator generates user utterances based on dialogue history and expresses doubt when hallucination is detected using a hallucination-aware prompt.

We define two user types. \emph{Action-oriented} users regulate emotions through actions and may not immediately articulate causes, while \emph{Emotion-oriented} users require emotional validation before accepting practical suggestions. The simulator may terminate the dialogue when it deems the interaction complete. Appendix~\ref{app:user_sim_prompt} provides user simulation prompts and type definitions.

\paragraph{Episode Generation.}
Each episode begins with an initial user utterance. Agent and user then alternate turns. Within each agent turn, the model may invoke tools multiple times before responding. Dialogue length is capped at 15 turns unless the simulator terminates earlier.

\begin{table*}[ht]
\centering
\begin{adjustbox}{width=1\textwidth}
\begin{tabular}{lccccccccc}
\toprule
\multirow{2.5}{*}{\textbf{Model}}
& \multicolumn{5}{c}{\textbf{TEA-Scores}}
& \multirow{2.5}{*}{\textbf{AVG.(TEA)} $\uparrow$}
& \multicolumn{3}{c}{\textbf{Factuality}} \\
\cmidrule(lr){2-6} \cmidrule(lr){8-10}

& Div. $\uparrow$ & Flu. $\uparrow$ & Hum. $\uparrow$ & Info. $\uparrow$ & Eff. $\uparrow$
& 
& Fact. $\uparrow$ & Halluc. $\downarrow$ & \textbf{Halluc. Rate} $\downarrow$ \\ 
\midrule

\rowcolor{gray!8}
\multicolumn{10}{c}{\emph{without tool}} \\
\midrule

\texttt{GPT-4o-mini} & 65.90  & 86.88  & 91.36  & \textbf{63.27}  & 75.93  & 76.67  & \textbf{82.38}  & 21.60  & 24.95   \\ 
\texttt{GPT-4.1-nano} & \underline{69.14}  & \textbf{90.74}  & \textbf{95.68}  & \textbf{63.27}  & \textbf{81.94}  & \textbf{80.15}  & 62.21  & \underline{11.35}  & \underline{14.51}   \\ 
\texttt{Gemini-2.5-flash} & 58.33  & 80.09  & 78.55  & 57.10  & 59.57  & 66.73  & 75.71  & 54.59  & 64.92   \\ 
\texttt{Qwen-plus} & 66.05  & 83.49  & 85.03  & 60.03  & 68.83  & 72.69  & 75.98  & 57.26  & 68.81   \\ 
\texttt{Qwen3-235B-a22B} & 67.13  & 84.26  & 83.64  & 59.26  & 65.12  & 71.88  & \underline{79.48}  & 61.05  & 71.21   \\ 
\texttt{Qwen3-next-80B-a3B} & \textbf{69.60}  & 86.42  & 88.89  & \underline{60.34}  & 80.09  & 77.07  & 64.44  & 41.85  & 62.49   \\ 
\texttt{Qwen3-32B} & 64.81  & 85.49  & 89.04  & 59.10  & 73.77  & 74.44  & 69.23  & 37.58  & 50.60   \\ 
\texttt{Qwen3-14B} & 64.66  & 87.35  & 92.44  & 58.95  & 80.71  & 76.82  & 59.81  & 15.68  & 21.76   \\ 
\texttt{Qwen3-8B} & 65.90  & \underline{87.96}  & \underline{93.98}  & 57.56  & \underline{81.79}  & \underline{77.44}  & 52.28  & \textbf{8.78}  & \textbf{11.41}   \\ 

\midrule
\rowcolor{gray!8}
\multicolumn{10}{c}{\emph{with tool}} \\
\midrule

\texttt{GPT-4o-mini} & \cellcolor{blue1!9} \underline{70.22} & \cellcolor{blue1!8} \textbf{91.05} & \cellcolor{blue1!8} \underline{95.52} & \cellcolor{blue1!10} \textbf{68.06} & \cellcolor{blue1!10} 80.71 & \cellcolor{blue1!9} \underline{81.11} & \cellcolor{red1!5} 77.61 & \cellcolor{blue1!7} 14.72 & \cellcolor{blue1!7} 18.44  \\ 
        \texttt{GPT-4.1-nano} & \cellcolor{blue1!2} \underline{70.22} & \cellcolor{red1!1} \underline{90.43} & \cellcolor{blue1!2} \textbf{96.91} & \cellcolor{blue1!2} 64.2 & \cellcolor{blue1!5} \textbf{84.26} & \cellcolor{blue1!2} \textbf{81.2} & \cellcolor{red1!5} 57.6 & \cellcolor{blue1!3} 8.64 & \cellcolor{blue1!3} \underline{11.69}  \\ 
        \texttt{Gemini-2.5-flash} & \cellcolor{blue1!15} 66.05 & \cellcolor{blue1!18} 89.04 & \cellcolor{blue1!26} 91.51 & \cellcolor{blue1!9} 61.57 & \cellcolor{blue1!40} 79.48 & \cellcolor{blue1!22} 77.53 & \cellcolor{red1!10} 65.74 & \cellcolor{blue1!37} 17.25 & \cellcolor{blue1!44} 21.01  \\ 
        \texttt{Qwen-plus} & \cellcolor{blue1!6} 68.83 & \cellcolor{blue1!11} 88.89 & \cellcolor{blue1!10} 90.12 & \cellcolor{blue1!8} 63.89 & \cellcolor{blue1!20} 78.7 & \cellcolor{blue1!11} 78.09 & \cellcolor{blue1!7} \textbf{83.24} & \cellcolor{blue1!27} 30.05 & \cellcolor{blue1!34} 34.89  \\ 
        \texttt{Qwen3-235B-a22B} & \cellcolor{blue1!4} 68.98 & \cellcolor{blue1!11} 89.51 & \cellcolor{blue1!16} 91.67 & \cellcolor{blue1!12} \underline{65.43} & \cellcolor{blue1!32} 81.02 & \cellcolor{blue1!15} 79.32 & \cellcolor{red1!1} \underline{78.42} & \cellcolor{blue1!36} 25.5 & \cellcolor{blue1!40} 31.44  \\ 
        \texttt{Qwen3-next-80B-a3B} & \cellcolor{blue1!4} \textbf{71.76} & \cellcolor{blue1!7} 89.97 & \cellcolor{blue1!5} 91.2 & \cellcolor{blue1!2} 61.57 & \cellcolor{red1!2} 79.17 & \cellcolor{blue1!3} 78.73 & \cellcolor{blue1!5} 69.44 & \cellcolor{blue1!1} 40.85 & \cellcolor{blue1!5} 57.57  \\ 
        \texttt{Qwen3-32B} & \cellcolor{blue1!6} 67.59 & \cellcolor{blue1!8} 89.51 & \cellcolor{blue1!11} 94.29 & \cellcolor{blue1!1} 59.57 & \cellcolor{blue1!21} \underline{84.1} & \cellcolor{blue1!9} 79.01 & \cellcolor{red1!1} 68.21 & \cellcolor{blue1!13} 24.13 & \cellcolor{blue1!17} 34.03  \\ 
        \texttt{Qwen3-14B} & \cellcolor{red1!2} 63.58 & \cellcolor{blue1!0} 87.35 & \cellcolor{blue1!5} 94.75 & \cellcolor{red1!6} 56.02 & \cellcolor{red1!1} 80.09 & \cellcolor{red1!1} 76.36 & \cellcolor{red1!6} 53.89 & \cellcolor{blue1!7} \underline{8.57} & \cellcolor{blue1!9} 12.45  \\ 
        \texttt{Qwen3-8B} & \cellcolor{blue1!0} 65.9 & \cellcolor{red1!0} 87.81 & \cellcolor{blue1!3} 95.37 & \cellcolor{blue1!1} 58.18 & \cellcolor{red1!0} 81.64 & \cellcolor{blue1!1} 77.78 & \cellcolor{red1!0} 51.95 & \cellcolor{blue1!3} \textbf{5.41} & \cellcolor{blue1!4} \textbf{7.16}  \\ 
\bottomrule
\end{tabular}
\end{adjustbox}
\caption{
Evaluation on \texttt{TEA-Bench}. For the \emph{with tool} setting, \colorbox{blue1!30}{blue} cells indicate improvement compared to \emph{without tool}, while \colorbox{red1!30}{red} cells indicate performance drop. The intensity of the color reflects the magnitude of the change. \textbf{Bold} and \underline{underlined} numbers denote the best and second-best results, respectively.
}
\label{tab:main_result}
\vspace{-0.5cm}
\end{table*}

\subsection{Evaluation Metrics}
\label{sec:evaluation_metrics}

\texttt{TEA-Bench} evaluates agents at the dialogue level, jointly considering empathetic support quality and factual grounding. As shown in Figure~\ref{fig:main}(b), we use a language model as an automatic evaluator to score complete dialogue episodes.

\paragraph{TEA-Scores.}
We assess empathetic support quality using five key dimensions. Four are directly adapted from ESC-Eval \citep{zhao2024esc}: \textbf{Diversity}, \textbf{Fluency}, \textbf{Humanoid}, and \textbf{Information}. We introduce the fifth dimension, \textbf{Effectiveness}, which measures whether the agent's suggestions are accepted and integrated by the user.

The first three dimensions assess dialogue quality and the agent's affective support, while the latter two evaluate instrumental support through the quality and relevance of recommendation. Each dimension receives an integer score from 0 to 4 based on the full dialogue. Raw scores are normalized to 0-100 and averaged to produce the TEA score. Appendix~\ref{app:details_of_tea_metrics} provides detailed rubrics and Appendix~\ref{app:eval_prompt} shows the evaluation prompt.

\paragraph{Factuality Metrics.}
We quantify factual grounding using three dialogue-level statistics: (i)~\textbf{factual content ratio}, the proportion of agent responses containing factual claims; (ii)~\textbf{hallucination ratio}, the proportion of responses containing hallucinated content; (iii)~\textbf{hallucination rate}, the ratio of hallucinated content to all factual content.

A factual claim is considered hallucinated if it cannot be traced to user-provided information or tool observations (Section~\ref{sec:interactive_framework}). All metrics are averaged across dialogues, ensuring that each interaction contributes equally. Together, these metrics characterize how frequently agents include factual information and how reliably it is grounded. Appendix~\ref{app:details_of_halluc_metrics} provides detailed definitions.

\section{Experiments}

\subsection{Experimental Setup}
\label{sec:exp_setup}

We evaluate various LLMs under TEA-bench, including four closed-source models \emph{GPT‑4o-mini}~\citep{hurst2024gpt}, \emph{GPT-4.1-nano}~\citep{openai2025gpt41}, \emph{Gemini-2.5-flash}~\citep{comanici2025gemini} and \emph{Qwen-plus}~\citep{yang2025qwen3}, as well as five open-source models \emph{Qwen3‑-235B-‑A22B-‑Instruct-2507}~\citep{yang2025qwen3}, \emph{Qwen3-Next-80B-A3B-Instruct}~\citep{yang2025qwen3}, \emph{Qwen3-32B}~\citep{yang2025qwen3}, \emph{Qwen3‑-14B}~\citep{yang2025qwen3} and \emph{Qwen3-‑8B}~\citep{yang2025qwen3}.

\subsection{RQ1: How Well Do Current Models Perform in Tool-enhanced ESC?}
\label{sec:rq1}

This research question examines whether tool augmentation improves ESC performance. We evaluate nine representative models under two settings: \emph{without tool} and \emph{with tool}. The main results are reported in Table~\ref{tab:main_result}. The reliability of these automatic metrics is validated through human evaluation, as detailed in Appendix~\ref{app:human_eval}.

\paragraph{Overall Impact of Tool Augmentation.}
Overall, enabling tool use improves \texttt{TEA-Scores} for most evaluated models. Models with strong reasoning and tool-calling capabilities exhibit substantial gains, indicating that tools provide useful contextual grounding for empathetic responses. In contrast, models with weaker tool usage ability may experience limited improvement or slight degradation, suggesting that effective tool integration is a prerequisite for performance gains.

\paragraph{Hallucination and Factual Grounding.}
Across all models, tool augmentation consistently reduces hallucination-related metrics. Both the proportion of hallucinated factual content and the hallucination rate decrease under the \emph{with tool} setting, demonstrating that tools reliably enhance factual grounding even when empathy gains are modest.

\paragraph{Results Across User Types.}
We further report results under different user types in Appendix~\ref{app:user_type_performance}. Tool augmentation yields particularly strong improvements for action-oriented users, while performance for emotion-oriented users improves for most models but may decline slightly for weaker ones. Importantly, hallucination rates are consistently reduced for both user types.

These results demonstrate that tool augmentation can effectively enhance overall ESC performance and factual grounding, motivating further analysis of tool usage behaviors in RQ2.

\begin{figure}[t]
    \centering
    \includegraphics[width=\linewidth]{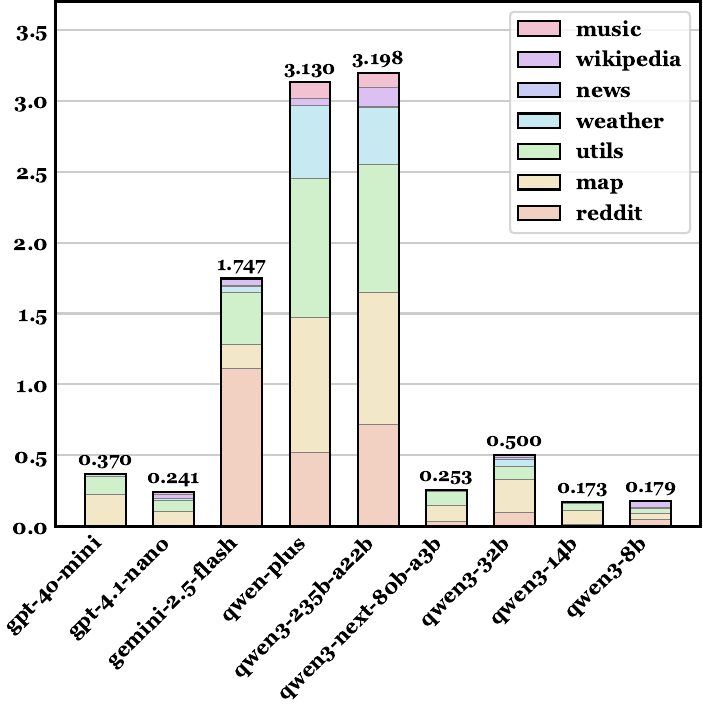}
    \caption{
Average number of tool calls per dialogue across different models on \texttt{TEA-Bench}.
}
    \label{fig:tool_usage}
    \vspace{-0.3cm}
\end{figure}

\begin{figure}[t]
    \centering
    \includegraphics[width=\linewidth]{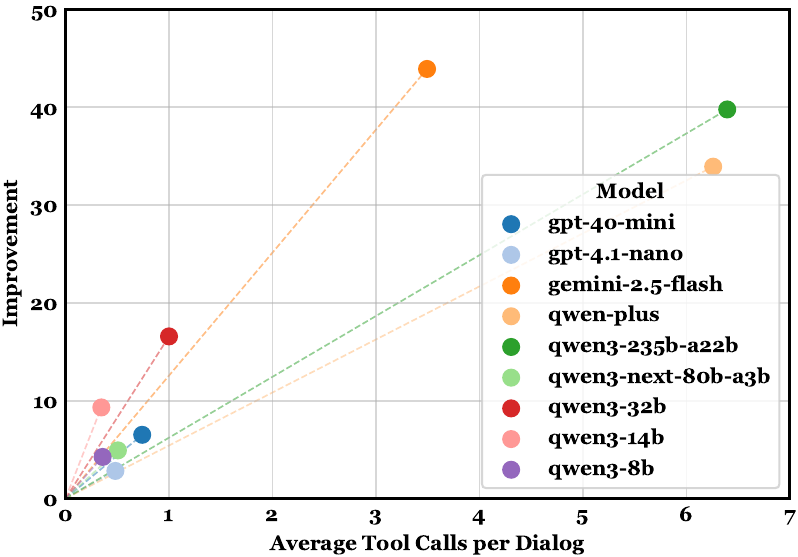}
    \caption{
Relationship between tool usage frequency and hallucination rate reduction across models. Dashed lines to the origin indicate per-call efficiency, with steeper slopes corresponding to higher efficiency.
    }
    \label{fig:usage_reduction}
    \vspace{-0.5cm}
\end{figure}

\begin{figure}[t]
    \centering
    \includegraphics[width=\linewidth]{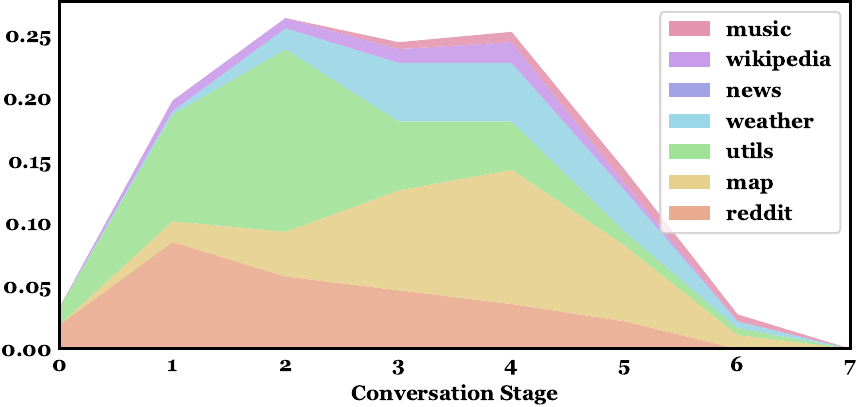}
    \caption{Average tool usage distribution across normalized dialogue stages in \texttt{TEA-Dialog}. Different colors represent different tool categories.}
    \label{fig:dataset_stage_avg}
    \vspace{-0.5cm}
\end{figure}

\begin{table*}[ht]
\centering
\begin{adjustbox}{width=1\textwidth}
\begin{tabular}{lccccccccc}
\toprule
\multirow{2.5}{*}{\textbf{Model}}
& \multicolumn{5}{c}{\textbf{TEA-Bench}}
& \multirow{2.5}{*}{\textbf{AVG.(TEA)} $\uparrow$}
& \multicolumn{3}{c}{\textbf{Hallucination}} \\
\cmidrule(lr){2-6} \cmidrule(lr){8-10}

& Div. $\uparrow$ & Flu. $\uparrow$ & Hum. $\uparrow$ & Info. $\uparrow$ & Eff. $\uparrow$
& 
& Fact. $\uparrow$ & Halluc. $\downarrow$ & \textbf{Halluc. Rate} $\downarrow$ \\ 
\midrule

\rowcolor{gray!8}
\multicolumn{10}{c}{\emph{Qwen3-8B}} \\
\midrule

\texttt{Base} & 65.9 & 87.81 & 95.37 & 58.18 & 81.64 & 77.78 & 51.95 & 5.41 & 7.16  \\ 
        \texttt{\textbf{TEA}} & \cellcolor{blue1!8} 69.91 & \cellcolor{red1!1} 87.19 & \cellcolor{red1!6} 92.59 & \cellcolor{blue1!4} 60.34 & \cellcolor{blue1!2} 82.56 & \cellcolor{blue1!1} 78.52 & \cellcolor{blue1!17} 68.65 & \cellcolor{red1!1} 6.86 & \cellcolor{red1!3} 9.73  \\ 
        ~~~\texttt{TEA-ID} & \cellcolor{blue1!12} 71.67 & \cellcolor{blue1!1} 88.54 & \cellcolor{red1!2} 94.17 & \cellcolor{blue1!9} 62.71 & \cellcolor{blue1!2} 82.5 & \cellcolor{blue1!4} 79.92 & \cellcolor{blue1!18} 70.38 & \cellcolor{blue1!2} 3.73 & \cellcolor{blue1!3} 4.61  \\ 
        ~~~\texttt{TEA-OOD} & \cellcolor{red1!2} 64.88 & \cellcolor{red1!9} 83.33 & \cellcolor{red1!15} 88.1 & \cellcolor{red1!9} 53.57 & \cellcolor{blue1!2} 82.74 & \cellcolor{red1!7} 74.52 & \cellcolor{blue1!12} 63.71 & \cellcolor{red1!10} 15.8 & \cellcolor{red1!17} 24.39  \\

\midrule
\rowcolor{gray!8}
\multicolumn{10}{c}{\emph{Qwen3-14B}} \\
\midrule

        \texttt{Base} & 63.58 & 87.35 & 94.75 & 56.02 & 80.09 & 76.36 & 53.89 & 8.57 & 12.45  \\ 
        \texttt{\textbf{TEA}} & \cellcolor{blue1!7} 66.98 & \cellcolor{red1!1} 86.88 & \cellcolor{red1!3} 93.36 & \cellcolor{blue1!9} 60.34 & \cellcolor{blue1!5} 82.41 & \cellcolor{blue1!3} 77.99 & \cellcolor{blue1!12} 66.32 & \cellcolor{red1!1} 9.12 & \cellcolor{blue1!0} 12.2  \\ 
        ~~~\texttt{TEA-ID} & \cellcolor{blue1!9} 67.92 & \cellcolor{red1!1} 86.88 & \cellcolor{red1!1} 94.17 & \cellcolor{blue1!9} 60.42 & \cellcolor{blue1!4} 82.29 & \cellcolor{blue1!4} 78.34 & \cellcolor{blue1!14} 68.09 & \cellcolor{blue1!1} 7.4 & \cellcolor{blue1!3} 9.86  \\ 
        ~~~\texttt{TEA-OOD} & \cellcolor{blue1!1} 64.29 & \cellcolor{red1!1} 86.9 & \cellcolor{red1!7} 91.07 & \cellcolor{blue1!8} 60.12 & \cellcolor{blue1!5} 82.74 & \cellcolor{blue1!1} 77.02 & \cellcolor{blue1!7} 61.25 & \cellcolor{red1!5} 14.03 & \cellcolor{red1!6} 18.89  \\

\bottomrule
\end{tabular}
\end{adjustbox}
\caption{
Evaluation of TEA-trained models on in-domain (ID) and out-of-domain (OOD) scenarios. \colorbox{blue1!30}{blue} cells indicate improvement compared to \texttt{Base}, while \colorbox{red1!30}{red} cells indicate performance drop. The intensity of the color reflects the magnitude of the change.
}
\label{tab:method_result}
\vspace{-0.5cm}
\end{table*}

\subsection{RQ2: How Do Tools Improve ESC?}
\label{sec:rq2}

This research question investigates how tools contribute to improved ESC performance and hallucination mitigation. We analyze tool usage behaviors across models, their relationship with hallucination reduction, and the characteristics of high-quality tool-enhanced dialogues.

\paragraph{Tool Usage Frequency across Models.}
Figure~\ref{fig:tool_usage} compares the average number of tool calls per dialogue across different models. Stronger models achieve substantial performance gains with relatively fewer tool calls, indicating more precise and effective tool usage. In contrast, mid-capability models rely on more frequent tool invocations to obtain comparable improvements, while weaker models invoke tools infrequently and show limited benefits, reflecting difficulties in effectively leveraging external tools. We further report results under different user types in Appendix~\ref{app:user_type_tool_usage}.

\paragraph{Tool Usage and Hallucination Reduction.}
We further examine the relationship between tool usage frequency and hallucination mitigation. As shown in Figure~\ref{fig:usage_reduction}, the reduction in hallucination rate generally increases with the number of tool calls, revealing a clear positive correlation between tool usage and hallucination elimination. However, models vary substantially in efficiency: stronger models achieve larger hallucination reductions with fewer tool calls, whereas mid-capability models require more frequent interactions to obtain similar gains. Smaller models exhibit limited absolute hallucination reduction despite occasional high per-call efficiency.

\paragraph{\texttt{TEA-Dialog}: High-quality Tool-enhanced Dialogue Dataset.}
To better understand effective tool usage patterns, we collect \texttt{TEA-Dialog}, a dataset of 365 high-quality dialogues. Dialogues are selected from multiple models based on high \texttt{TEA-Scores} (above 80) and absence of detected hallucinations, with additional human filtering to ensure reliability.

Figure~\ref{fig:dataset_stage_avg} shows the average tool usage across normalized dialogue stages. Early stages primarily involve utility and contextual tools to establish situational grounding, followed by environment-aware tools such as \texttt{maps} and \texttt{weather}. In later stages feature personalized tools, including \texttt{music} and \texttt{news}, reflecting a structured transition from context acquisition to tailored emotional support. Additional analyses on tool usage patterns under different user types and other dialogue attributes are provided in Appendix~\ref{app:dataset_analysis}.

\subsection{RQ3: How Can Tool-enhanced ESC Be Further Improved?}
\label{sec:rq3}

We investigate whether supervised fine-tuning (SFT) on \texttt{TEA-Dialog} can further improve ESC performance. Using dialogues from the first 60 scenarios, we fine-tune \texttt{Qwen3-8B} and \texttt{Qwen3-14B}, yielding \texttt{Qwen3-8B-TEA} and \texttt{Qwen3-14B-TEA}. Models are evaluated on all 81 scenarios, divided into \textbf{In-Domain (ID)} scenarios seen during training and \textbf{Out-Of-Domain (OOD)} scenarios with unseen descriptions. Training details are in Appendix~\ref{app:training_details}.

\paragraph{Overall Performance.}
Both TEA-trained models achieve consistent improvements over their base counterparts across \texttt{TEA-Scores}, particularly \emph{Information} and \emph{Effectiveness}. These gains are accompanied by a substantial increase in factual content, indicating that SFT encourages models to produce more grounded responses.

\paragraph{Generalization under Distribution Shift.}
Performance on ID scenarios is significantly higher than on OOD scenarios for both models, suggesting limited generalization when training on a small set of high-quality scenarios. Notably, \texttt{Qwen3-14B} exhibits a smaller gap between ID and OOD than \texttt{Qwen3-8B}, indicating that larger models generalize better at the scenario level.

\paragraph{Hallucination Trade-offs.}
While hallucination rates decrease on ID scenarios after fine-tuning, they increase markedly on OOD scenarios. This trend suggests that although SFT improves factual grounding in familiar contexts, it may also amplify hallucination risks under distribution shift.

These results indicate that naïve SFT on limited high-quality data is insufficient for robustly improving tool-enhanced ESC, motivating the need for more effective training or alignment strategies.

\section{Conclusion}

We emphasize ESC with instrumental support to ensure trustworthiness and actionable guidance. Existing systems focus on affective expression in text-only settings, leaving factual grounding and hallucinations underexplored. We introduce \texttt{TEA-Bench}, a benchmark for tool-augmented ESC agents, and \texttt{TEA-Dialog}, a dataset of tool-enhanced dialogues. Experiments across nine LLMs show tool augmentation improves support quality and reduces hallucination, with gains dependent on model capacity. Fine-tuning on \texttt{TEA-Dialog} boosts in-distribution performance but generalizes poorly, highlighting challenges of reliable, tool-augmented ESC agents.

\section*{Limitations}

\paragraph{Simulated User.}
\texttt{TEA-Bench} relies on a simulated user to enable controlled, reproducible, and scalable evaluation. While this allows for fine-grained and process-level assessment, it cannot fully capture the diversity, spontaneity, and unpredictability of real human behavior in ESC.

\paragraph{Interaction Horizon.}
Our evaluation focuses on short to medium-length ESC interactions. Long-term dynamics such as sustained emotional support, trust formation, and user adaptation across extended conversations are not explicitly modeled and remain an important direction for future work.

\paragraph{Generalization of Training Data.}
Although fine-tuning on \texttt{TEA-Dialog} improves performance in familiar scenarios, we observe limited generalization under distribution shift, and in some cases increased hallucination. This suggests that supervised fine-tuning alone may be insufficient for training robust and broadly generalizable tool-augmented emotional support agents.

\paragraph{Long-term Memory.} TEA-Bench focuses on immediate situational grounding via tools rather than long-term conversational memory. Recent benchmarks \citep{maharana2024evaluating, guo2026personalization, jiang2025know, hu2026op} have systematically evaluated memory recall and retention across extended dialogues, which remains unexplored in our work.

\section*{Ethical Statement}

We are committed to publicly releasing all data upon acceptance of the paper. We are fully aware of the potential biases associated with LLM-as-Judge. To mitigate these effects, we incorporated human expert assessments. However, due to cost considerations, the scale of human evaluation remains limited at this stage. We note that this constraint is common in current conversational AI research that relies on LLMs.

\section*{Acknowledgments}
We thank the anonymous reviewers for their comments and suggestions. This work was supported by the National Natural Science Foundation of China (NSFC) via grant 62441614 and 62576125.

\bibliography{custom}

\appendix

\section{Tool Categories and Data Sources}
\label{app:tool_sources}

Table~\ref{tab:tool_overview} summarizes the concrete contents provided by each tool category. News-related tools are backed by the GDELT Project, providing access to global news events. Weather information is obtained from Meteostat, while music-related queries rely on MusicBrainz metadata. Map information is supported by OpenStreetMap, and community discussions are retrieved from Reddit. Encyclopedic knowledge is accessed via Wikipedia. Within the \texttt{Utils} category, \texttt{utils\_get\_current\_time} and
\texttt{utils\_get\_user\_location} read predefined scenario attributes, while
\texttt{utils\_fetch\_webpage\_content} extracts plain text from a specified webpage
using \texttt{trafilatura}.

\begin{table}[htbp]
    \centering
    \setlength{\extrarowheight}{0pt}
    \renewcommand{\arraystretch}{1.2}
    \resizebox{\linewidth}{!}{
        \begin{tabular}{c c}
            \toprule
            \multirow{5}{*}{\texttt{Wikipedia}}
            & wikipedia\_search\_topics\\
            & wikipedia\_get\_summary\\
            & wikipedia\_get\_section\\
            & wikipedia\_get\_available\_sections\\
            & wikipedia\_get\_full\_content\\

            \midrule
            \multirow{4}{*}{\texttt{Map}}
            & map\_find\_route\\
            & map\_find\_nearby\_places\\
            & map\_calculate\_reachable\_area\\
            & map\_get\_location\_info\\

            \midrule
            \multirow{2}{*}{\texttt{Weather}}
            & weather\_get\_current\\
            & weather\_get\_forecast\\

            \midrule
            \multirow{3}{*}{\texttt{Utils}}
            & utils\_get\_current\_time\\
            & utils\_get\_user\_location\\
            & utils\_fetch\_webpage\_content\\

            \midrule
            \multirow{4}{*}{\texttt{News}}
            & news\_search\_by\_keywords\\
            & news\_get\_related\_themes\\
            & news\_search\_by\_theme\\
            & news\_search\_by\_location\\

            \midrule
            \multirow{7}{*}{\texttt{Music}}
            & music\_search\_artists\\
            & music\_search\_releases\\
            & music\_search\_recordings\\
            & music\_search\_releases\_by\_year\\
            & music\_get\_artist\_details\\
            & music\_get\_release\_details\\
            & music\_get\_recording\_details\\

            \midrule
            \multirow{6}{*}{\texttt{Reddit}}
            & reddit\_search\_posts\\
            & reddit\_search\_subreddit\\
            & reddit\_get\_subreddit\_posts\\
            & reddit\_get\_post\_comments\\
            & reddit\_get\_subreddit\_info\\
            & reddit\_search\_subreddits\\
            
            \bottomrule
        \end{tabular}
    }
    \caption{The tools and their corresponding category}
    \label{tab:tool_overview}
\end{table}

\section{Details of Metrics}
\label{app:details_of_metrics}

\subsection{TEA Metrics.}
\label{app:details_of_tea_metrics}

The explanations of each metric are as follows:

\paragraph{Diversity (Div.)} Focusing on the diversity of expression forms and the richness of content in dialogue.

\paragraph{Fluency (Flu.)} Not only focus on the logical coherence of the context in dialogues but also pay attention to the fluency of expression in a given conversation.

\paragraph{Humanoid (Hum.)} Focus on the differences between emotional assistants and humans.

\paragraph{Information (Info.)} Focusing on Evaluating the Reasonableness and Quantity of Recommendations Provided by Emotion Assistants.

\paragraph{Effectiveness (Eff.)} Focusing on whether the supporter’s suggestions are practically accepted and meaningfully integrated by the help-seeker, as reflected in their subsequent responses or emotional shifts.

Evaluation rules are listed in Table \ref{tab:evaluation_criteria}.

\begin{table*}
    \centering
    \small
    \renewcommand{\arraystretch}{1.8}
    \resizebox{\linewidth}{!}{
    \begin{tabularx}{\textwidth}{|c|X|X|X|X|X|X|}
        \hline
\centering{\textbf{Score}} & \centering{\textbf{Diversity}} & \centering{\textbf{Fluency}} & \centering{\textbf{Humanoid}} & \centering{\textbf{Information}} & \multicolumn{1}{c|}{\textbf{Effectiveness}}  \\ 
        \hline
        \textbf{0 points}  & The dialogue exhibits rigidity and lacks comprehension in terms of internalizing the content. & There are significant issues with comprehending the content, logic, and expression in the dialogue, rendering it completely incomprehensible. & The dialogue exhibits rigidity and lacks comprehension in terms of internalizing the content. & Suggestions were provided, but all of them were ineffective, and some even gave advice that could potentially harm the user. & The suggestions are invalidating, harmful, or coercive.  \\ 
        \hline
        \textbf{1 point} & The expression form is monotonous and lacks substantive content. & The content of the dialogue can be understood to some extent, although there are certain issues with the logic and expression employed. & Structured responses, or responses in the form of ’As a large language model’ or robot-like replies. & Have suggestions but ineffective, as well as no suggestions. & The suggestions are inappropriate or minimally useful in context.  \\ 
        \hline
        \textbf{2 points} & The expression form is monotonous or lacks substantive content. & The dialogue exhibits good readability in terms of content, but there are issues with either the logical coherence or the expression employed. & More than two traces can reveal that the AI assistant is a language model. & The suggestions are fewer than five, and some suggestions are effective, while others provide numerous suggestions, but none of them touch the root of the problem. & The suggestions are partially applicable but lack personalization or timing; they may be generic, overly simplistic, or mismatched to the help-seeker’s current state.  \\ 
        \hline
        \textbf{3 points} & The dialogue content demonstrates a high level of readability without any apparent issues. & The dialogue content demonstrates a high level of readability without any apparent issues. & 1-2 traces can reveal that the AI assistant is a language model. & There are more than five suggestions, but some of them are ineffective. There are fewer than five suggestions, but all of them are very effective. & The suggestions are relevant and acknowledged without rejection, but there is no observable internalization or change in the help-seeker’s tone, stance, or intent.  \\ 
        \hline
        \textbf{4 points} & The form exhibits diversity, while demonstrating a high degree of content richness. & The dialogue content exhibits a high level of readability, comprehensive logical coherence, and outstanding expression. & There is no apparent difference from human friends. & There are many suggestions, and all of them are effective. & The suggestions are not only reasonable but are actively embraced by the help-seeker, leading to clear emotional resonance or behavioral commitment.  \\ 
        \hline
    \end{tabularx}}
    \caption{Evaluation criteria of \texttt{TEA} Metrics.}
    \label{tab:evaluation_criteria}
    
\end{table*}

\subsection{Factuality and Hallucination Metrics.}
\label{app:details_of_halluc_metrics}

Let a dialogue $d$ consist of a sequence of model responses $\{r_1, r_2, \dots, r_{T_d}\}$. For each response, we annotate whether it contains factual content and whether such content is hallucinated according to the grounding criteria described in Section~\ref{sec:interactive_framework}.

For each dialogue $d$, we define:
\begin{itemize}
    \item $T_d$: the total number of model responses in the dialogue;
    \item $F_d$: the number of responses containing factual content;
    \item $H_d$: the number of responses containing hallucinated factual content.
\end{itemize}

Using these quantities, we compute the following dialogue-level metrics:
\begin{equation}
\text{Fact.}_d = \frac{F_d}{T_d},
\end{equation}
\begin{equation}
\text{Halluc.}_d = \frac{H_d}{T_d},
\end{equation}
\begin{equation}
\text{Halluc. Rate}_d = \frac{H_d}{F_d}, \quad \text{where } F_d > 0.
\end{equation}

The final reported scores are obtained by macro-averaging across all dialogues:
\begin{equation}
\text{Fact.} = \frac{1}{|\mathcal{D}|} \sum_{d \in \mathcal{D}} \text{Fact.}_d,
\end{equation}
\begin{equation}
\text{Halluc.} = \frac{1}{|\mathcal{D}|} \sum_{d \in \mathcal{D}} \text{Halluc.}_d,
\end{equation}
\begin{equation}
\text{Halluc. Rate} = \frac{1}{|\mathcal{D}'|} \sum_{d \in \mathcal{D}'} \text{Halluc. Rate}_d,
\end{equation}
where $\mathcal{D}$ denotes the full set of evaluated dialogues and $\mathcal{D}' \subseteq \mathcal{D}$ includes only dialogues with $F_d > 0$.

This dialogue-level aggregation ensures that each interaction contributes equally to the final evaluation, mitigating biases introduced by dialogue length or factual density and better reflecting interaction-level grounding behavior in emotional support conversations.

\begin{table*}[ht]
\centering
\begin{adjustbox}{width=1\textwidth}
\begin{tabular}{lccccccccc}
\toprule
\multirow{2.5}{*}{\textbf{Model}}
& \multicolumn{5}{c}{\textbf{TEA-Scores}}
& \multirow{2.5}{*}{\textbf{AVG.(TEA)} $\uparrow$}
& \multicolumn{3}{c}{\textbf{Factuality}} \\
\cmidrule(lr){2-6} \cmidrule(lr){8-10}

& Div. $\uparrow$ & Flu. $\uparrow$ & Hum. $\uparrow$ & Info. $\uparrow$ & Eff. $\uparrow$
& 
& Fact. $\uparrow$ & Halluc. $\downarrow$ & \textbf{Halluc. Rate} $\downarrow$ \\ 
\midrule

\rowcolor{gray!8}
\multicolumn{10}{c}{\emph{without tool}} \\
\midrule

\texttt{GPT-4o-mini} & 70.37 & 91.98 & 93.83 & 71.91 & 85.49 & 82.72 & 90.05 & 28.06 & 29.99  \\ 
        \texttt{GPT-4.1-nano} & 70.99 & 94.44 & 94.44 & 71.6 & 86.73 & 83.64 & 77.44 & 17.1 & 19.11  \\ 
        \texttt{Gemini-2.5-flash} & 64.2 & 83.33 & 82.72 & 65.43 & 71.91 & 73.52 & 82.81 & 51.59 & 58.2  \\ 
        \texttt{Qwen-plus} & 66.05 & 82.41 & 83.33 & 63.89 & 66.67 & 72.47 & 87.09 & 66.7 & 72.51  \\ 
        \texttt{Qwen3-235B-a22B} & 67.59 & 84.26 & 83.33 & 62.65 & 65.74 & 72.71 & 88.45 & 65.73 & 70.57  \\ 
        \texttt{Qwen3-next-80B-a3B} & 69.14 & 85.49 & 88.58 & 63.89 & 77.78 & 76.98 & 77.2 & 47.35 & 61.18  \\ 
        \texttt{Qwen3-32B} & 65.74 & 85.8 & 88.58 & 66.98 & 73.15 & 76.05 & 84.31 & 42.1 & 46.36  \\ 
        \texttt{Qwen3-14B} & 66.36 & 89.81 & 91.05 & 68.21 & 85.19 & 80.12 & 75.58 & 24.82 & 29.92  \\ 
        \texttt{Qwen3-8B} & 69.44 & 91.67 & 95.68 & 67.9 & 87.96 & 82.53 & 66.65 & 11.72 & 13.05  \\ 

\midrule
\rowcolor{gray!8}
\multicolumn{10}{c}{\emph{with tool}} \\
\midrule

\texttt{GPT-4o-mini} & \cellcolor{blue1!10} 75.31 & \cellcolor{blue1!7} 95.68 & \cellcolor{blue1!8} 97.84 & \cellcolor{blue1!14} 78.7 & \cellcolor{blue1!12} 91.67 & \cellcolor{blue1!10} 87.84 & \cellcolor{red1!4} 86.46 & \cellcolor{blue1!12} 16.37 & \cellcolor{blue1!12} 18.05 \\ 
        \texttt{GPT-4.1-nano} & \cellcolor{blue1!6} 74.07 & \cellcolor{red1!3} 92.9 & \cellcolor{blue1!6} 97.22 & \cellcolor{blue1!9} 76.23 & \cellcolor{blue1!9} 91.36 & \cellcolor{blue1!5} 86.36 & \cellcolor{red1!2} 75.63 & \cellcolor{blue1!3} 14.17 & \cellcolor{blue1!2} 17.17 \\ 
        \texttt{Gemini-2.5-flash} & \cellcolor{blue1!14} 71.3 & \cellcolor{blue1!18} 92.28 & \cellcolor{blue1!24} 94.75 & \cellcolor{blue1!14} 72.22 & \cellcolor{blue1!36} 89.81 & \cellcolor{blue1!21} 84.07 & \cellcolor{red1!4} 79.15 & \cellcolor{blue1!33} 18.15 & \cellcolor{blue1!38} 20.65 \\ 
        \texttt{Qwen-plus} & \cellcolor{blue1!15} 73.77 & \cellcolor{blue1!20} 92.59 & \cellcolor{blue1!24} 95.37 & \cellcolor{blue1!17} 72.53 & \cellcolor{blue1!48} 90.43 & \cellcolor{blue1!25} 84.94 & \cellcolor{red1!1} 86.07 & \cellcolor{blue1!48} 18.33 & \cellcolor{blue1!52} 20.09 \\ 
        \texttt{Qwen3-235B-a22B} & \cellcolor{blue1!11} 72.84 & \cellcolor{blue1!19} 93.52 & \cellcolor{blue1!27} 96.6 & \cellcolor{blue1!21} 73.15 & \cellcolor{blue1!55} 93.21 & \cellcolor{blue1!26} 85.86 & \cellcolor{red1!9} 79.06 & \cellcolor{blue1!55} 11.01 & \cellcolor{blue1!58} 12.65 \\ 
        \texttt{Qwen3-next-80B-a3B} & \cellcolor{blue1!9} 73.46 & \cellcolor{blue1!12} 91.67 & \cellcolor{blue1!6} 91.36 & \cellcolor{blue1!9} 68.21 & \cellcolor{blue1!4} 79.94 & \cellcolor{blue1!8} 80.93 & \cellcolor{blue1!8} 85.5 & \cellcolor{blue1!2} 44.96 & \cellcolor{blue1!9} 51.79 \\ 
        \texttt{Qwen3-32B} & \cellcolor{blue1!9} 70.37 & \cellcolor{blue1!14} 92.9 & \cellcolor{blue1!17} 96.91 & \cellcolor{blue1!6} 69.75 & \cellcolor{blue1!34} 90.12 & \cellcolor{blue1!16} 84.01 & \cellcolor{red1!1} 83.68 & \cellcolor{blue1!19} 23.36 & \cellcolor{blue1!20} 26.22 \\ 
        \texttt{Qwen3-14B} & \cellcolor{blue1!4} 68.52 & \cellcolor{blue1!2} 90.74 & \cellcolor{blue1!9} 95.37 & \cellcolor{blue1!1} 68.52 & \cellcolor{blue1!3} 86.73 & \cellcolor{blue1!4} 81.98 & \cellcolor{red1!5} 70.93 & \cellcolor{blue1!13} 12.09 & \cellcolor{blue1!16} 14.04 \\ 
        \texttt{Qwen3-8B} & \cellcolor{blue1!3} 70.99 & \cellcolor{blue1!1} 92.28 & \cellcolor{blue1!2} 96.91 & \cellcolor{blue1!7} 71.3 & \cellcolor{blue1!3} 89.51 & \cellcolor{blue1!3} 84.2 & \cellcolor{blue1!7} 73.72 & \cellcolor{blue1!3} 9.2 & \cellcolor{blue1!2} 10.79 \\ 

\bottomrule
\end{tabular}
\end{adjustbox}
\caption{
Evaluation on \texttt{TEA-Bench} under Action-oriented scenarios. For the \emph{with tool} setting, \colorbox{blue1!30}{blue} cells indicate improvement compared to \emph{without tool}, while \colorbox{red1!30}{red} cells indicate performance drop. The intensity of the color reflects the magnitude of the change.
}
\label{tab:action_result}
\vspace{-0.4cm}
\end{table*}

\begin{table*}[ht]
\centering
\begin{adjustbox}{width=1\textwidth}
\begin{tabular}{lccccccccc}
\toprule
\multirow{2.5}{*}{\textbf{Model}}
& \multicolumn{5}{c}{\textbf{TEA-Scores}}
& \multirow{2.5}{*}{\textbf{AVG.(TEA)} $\uparrow$}
& \multicolumn{3}{c}{\textbf{Factuality}} \\
\cmidrule(lr){2-6} \cmidrule(lr){8-10}

& Div. $\uparrow$ & Flu. $\uparrow$ & Hum. $\uparrow$ & Info. $\uparrow$ & Eff. $\uparrow$
& 
& Fact. $\uparrow$ & Halluc. $\downarrow$ & \textbf{Halluc. Rate} $\downarrow$ \\ 
\midrule

\rowcolor{gray!8}
\multicolumn{10}{c}{\emph{without tool}} \\
\midrule

 \texttt{GPT-4o-mini} & 61.42 & 81.79 & 88.89 & 54.63 & 66.36 & 70.62 & 74.71 & 15.15 & 19.91  \\ 
        \texttt{GPT-4.1-nano} & 67.28 & 87.04 & 96.91 & 54.94 & 77.16 & 76.67 & 46.99 & 5.6 & 9.91  \\ 
        \texttt{Gemini-2.5-flash} & 52.47 & 76.85 & 74.38 & 48.77 & 47.22 & 59.94 & 68.61 & 57.6 & 71.64  \\ 
        \texttt{Qwen-plus} & 66.05 & 84.57 & 86.73 & 56.17 & 70.99 & 72.9 & 64.87 & 47.82 & 65.11  \\ 
        \texttt{Qwen3-235B-a22B} & 66.67 & 84.26 & 83.95 & 55.86 & 64.51 & 71.05 & 70.51 & 56.37 & 71.85  \\ 
        \texttt{Qwen3-next-80B-a3B} & 70.06 & 87.35 & 89.2 & 56.79 & 82.41 & 77.16 & 51.68 & 36.35 & 63.8  \\ 
        \texttt{Qwen3-32B} & 63.89 & 85.19 & 89.51 & 51.23 & 74.38 & 72.84 & 54.15 & 33.07 & 54.85  \\ 
        \texttt{Qwen3-14B} & 62.96 & 84.88 & 93.83 & 49.69 & 76.23 & 73.52 & 44.05 & 6.55 & 13.59  \\ 
        \texttt{Qwen3-8B} & 62.35 & 84.26 & 92.28 & 47.22 & 75.62 & 72.35 & 37.91 & 5.85 & 9.76  \\ 

\midrule
\rowcolor{gray!8}
\multicolumn{10}{c}{\emph{with tool}} \\
\midrule

\texttt{GPT-4o-mini} & \cellcolor{blue1!7} 65.12 & \cellcolor{blue1!9} 86.42 & \cellcolor{blue1!9} 93.21 & \cellcolor{blue1!6} 57.41 & \cellcolor{blue1!7} 69.75 & \cellcolor{blue1!8} 74.38 & \cellcolor{red1!6} 68.76 & \cellcolor{blue1!2} 13.08 & \cellcolor{blue1!1} 18.84 \\ 
        \texttt{GPT-4.1-nano} & \cellcolor{red1!2} 66.36 & \cellcolor{blue1!2} 87.96 & \cellcolor{red1!1} 96.6 & \cellcolor{red1!6} 52.16 & \cellcolor{blue1!0} 77.16 & \cellcolor{red1!1} 76.05 & \cellcolor{red1!7} 39.56 & \cellcolor{blue1!2} 3.12 & \cellcolor{blue1!4} 6.21 \\ 
        \texttt{Gemini-2.5-flash} & \cellcolor{blue1!17} 60.8 & \cellcolor{blue1!18} 85.8 & \cellcolor{blue1!28} 88.27 & \cellcolor{blue1!4} 50.93 & \cellcolor{blue1!44} 69.14 & \cellcolor{blue1!22} 70.99 & \cellcolor{red1!16} 52.33 & \cellcolor{blue1!41} 16.35 & \cellcolor{blue1!50} 21.38 \\ 
        \texttt{Qwen-plus} & \cellcolor{red1!4} 63.89 & \cellcolor{blue1!1} 85.19 & \cellcolor{red1!4} 84.88 & \cellcolor{red1!2} 55.25 & \cellcolor{red1!8} 66.98 & \cellcolor{red1!3} 71.24 & \cellcolor{blue1!16} 80.42 & \cellcolor{blue1!6} 41.77 & \cellcolor{blue1!15} 49.68 \\ 
        \texttt{Qwen3-235B-a22B} & \cellcolor{red1!3} 65.12 & \cellcolor{blue1!2} 85.49 & \cellcolor{blue1!6} 86.73 & \cellcolor{blue1!4} 57.72 & \cellcolor{blue1!9} 68.83 & \cellcolor{blue1!3} 72.78 & \cellcolor{blue1!7} 77.78 & \cellcolor{blue1!16} 39.99 & \cellcolor{blue1!22} 50.24 \\ 
        \texttt{Qwen3-next-80B-a3B} & \cellcolor{blue1!0} 70.06 & \cellcolor{blue1!2} 88.27 & \cellcolor{blue1!4} 91.05 & \cellcolor{red1!4} 54.94 & \cellcolor{red1!8} 78.4 & \cellcolor{red1!1} 76.54 & \cellcolor{blue1!2} 53.39 & \cellcolor{red1!0} 36.74 & \cellcolor{blue1!0} 63.36 \\ 
        \texttt{Qwen3-32B} & \cellcolor{blue1!2} 64.81 & \cellcolor{blue1!2} 86.11 & \cellcolor{blue1!4} 91.67 & \cellcolor{red1!4} 49.38 & \cellcolor{blue1!7} 78.09 & \cellcolor{blue1!2} 74.01 & \cellcolor{red1!1} 52.73 & \cellcolor{blue1!8} 24.91 & \cellcolor{blue1!13} 41.83 \\ 
        \texttt{Qwen3-14B} & \cellcolor{red1!9} 58.64 & \cellcolor{red1!2} 83.95 & \cellcolor{blue1!1} 94.14 & \cellcolor{red1!12} 43.52 & \cellcolor{red1!6} 73.46 & \cellcolor{red1!6} 70.74 & \cellcolor{red1!7} 36.84 & \cellcolor{blue1!1} 5.06 & \cellcolor{blue1!3} 10.86 \\ 
        \texttt{Qwen3-8B} & \cellcolor{red1!3} 60.8 & \cellcolor{red1!2} 83.33 & \cellcolor{blue1!3} 93.83 & \cellcolor{red1!4} 45.06 & \cellcolor{red1!4} 73.77 & \cellcolor{red1!2} 71.36 & \cellcolor{red1!8} 30.19 & \cellcolor{blue1!4} 1.62 & \cellcolor{blue1!6} 3.54 \\ 
    
\bottomrule
\end{tabular}
\end{adjustbox}
\caption{
Evaluation on \texttt{TEA-Bench} under Emotion-oriented scenarios. For the \emph{with tool} setting, \colorbox{blue1!30}{blue} cells indicate improvement compared to \emph{without tool}, while \colorbox{red1!30}{red} cells indicate performance drop. The intensity of the color reflects the magnitude of the change.
}
\label{tab:emotion_result}
\vspace{-0.4cm}
\end{table*}

\section{Human Evaluation}
\label{app:human_eval}

\subsection{Human Evaluation of TEA-Scores}
Table~\ref{tab:human_score} reports the correlation between automatic \texttt{TEA-Scores} and human judgments. We randomly sample 150 complete dialogue episodes from the main experiments. Three human annotators each evaluate 50 dialogues using the same 0--4 integer scale as the automatic evaluator, covering five dimensions (Information, Humanoid, Diversity, Fluency, and Effectiveness). The average score across the five dimensions is reported as the overall \texttt{TEA-Scores}. Human annotators were instructed using exactly the same evaluation prompt as described in Section~\ref{app:eval_prompt}, ensuring full consistency between human and automatic TEA score assessments.

As shown in Table~\ref{tab:human_score}, automatic scores exhibit consistent positive correlations with human ratings across all dimensions. In particular, the overall TEA score achieves strong correlations under Spearman ($\rho=0.7448$), Pearson ($r=0.7563$), and Kendall ($\tau=0.6174$), indicating that the proposed automatic evaluation reliably reflects human perception of ESC quality. Among individual dimensions, Effectiveness and Humanoid show relatively higher agreement, while Information presents comparatively lower correlation, suggesting that factual adequacy is more challenging to assess automatically.

\begin{table}[t]
\centering
\begin{adjustbox}{width=0.48\textwidth}
\begin{tabular}{lcccccc}
\toprule
    & Div. & Flu. & Hum. & Info. & Eff. & AVG.(TEA) \\
\midrule
    Spearman & 0.5192 & 0.5356 & 0.5744 & 0.4123 & 0.6061 & 0.7448  \\ 
    Pearson & 0.5371 & 0.5152 & 0.5568 & 0.3671 & 0.6495 & 0.7563  \\ 
    Kendall & 0.4678 & 0.5119 & 0.5220 & 0.3932 & 0.5598 & 0.6174 \\ 
\bottomrule
\end{tabular}
\end{adjustbox}
\caption{
Correlation between automatic \texttt{TEA-Scores} and human judgments across different dimensions.
}
\label{tab:human_score}
\vspace{-0.4cm}
\end{table}

\subsection{Human Verification of Hallucination Detection}
Table~\ref{tab:human_hallu} summarizes the human verification results for factual content and hallucination detection. We randomly sample 150 assistant responses at the turn level, each containing a random reply from a random dialogue. Three annotators independently label whether the response contains factual content and whether such content is hallucinated, with each annotator assessing 50 samples. Due to the imbalance between grounded and hallucinated factual claims, we report not only Precision, Recall, and F1, but also Matthews Correlation Coefficient (MCC) and Cohen’s Kappa to provide a more robust evaluation. Annotators followed the identical hallucination detection prompt detailed in Section~\ref{app:hdm_prompt}, including the definition of factual content and hallucinated claims.

The results show that the hallucination detection module achieves high precision (0.8947) and strong agreement with human judgments (MCC=0.7056, Cohen’s Kappa=0.699), indicating that automatically detected hallucinations largely correspond to genuine ungrounded factual content. Overall, these findings confirm the reliability of both the automatic \texttt{TEA-Scores} and the hallucination detection framework used in TEA-Bench.

\begin{table}[t]
\centering
\begin{adjustbox}{width=0.48\textwidth}
\begin{tabular}{lcccccc}

\toprule
& Precision & Recall & F1 & MCC & Cohen\_Kappa  \\ 
\midrule
    Fact. & 0.9238 & 0.8661 & 0.894 & 0.6222 & 0.6179  \\ 
    Halluc. & 0.8947 & 0.7612 & 0.8226 & 0.7056 & 0.6990 \\ 
\bottomrule

\end{tabular}
\end{adjustbox}
\caption{
Human verification of automatic hallucination detection.
}
\label{tab:human_hallu}
\vspace{-0.4cm}
\end{table}

\subsection{Annotator Training and Costs}
All human annotators were carefully selected and trained prior to evaluation. Annotators were recruited from a commercial crowdsourcing platform and were adult participants (aged 18+), primarily based in English-speaking regions. No personally identifiable information was collected. Each annotator underwent a 2-hour training session where they reviewed detailed annotation guidelines, example dialogues, and scoring criteria for both \texttt{TEA-Scores} and hallucination detection. During training, annotators practiced on 20 additional dialogues not included in the main evaluation set and received feedback from the lead researcher to ensure consistent understanding of each scoring dimension.

For the main evaluation, each annotator assessed 50 dialogues for TEA scoring and 50 responses for hallucination verification, amounting to approximately 10 person-hours per annotator. Annotators were compensated at a rate of \$25 per hour, which is consistent with or above the local minimum wage in their country of residence and considered adequate for the required annotation effort. All annotators were informed of the purpose of the study and how their annotations would be used for research and evaluation, and they provided informed consent prior to participation. The annotation protocol involves no personal or sensitive data and was determined to be exempt from formal ethics review by the authors’ institutional review process.

\section{User Type Analysis}
\label{app:user_type_analysis}

\subsection{Performance Comparison Across User Types}
\label{app:user_type_performance}

This appendix presents a fine-grained analysis of model performance across different user types in the main experiment, focusing on \emph{Action-oriented} and \emph{Emotion-oriented} users.

\paragraph{Action-oriented Users.}
Table~\ref{tab:action_result} reports the results for Action-oriented scenarios. Compared to the \emph{without tool} setting, all evaluated models exhibit consistent improvements in overall scores when tools are enabled, with several models achieving gains exceeding 10 points. Improvements are observed across nearly all evaluation dimensions, with particularly substantial gains in Effectiveness and Information. This indicates that tool augmentation enables models to provide more effective and informative action guidance, which is crucial for users explicitly seeking actionable support.

In addition to performance gains, tool usage leads to a pronounced reduction in hallucination rates for all models. For several models, hallucinations are reduced by more than 50 points, suggesting that external tools play a critical role in grounding action-related recommendations and factual information. These results demonstrate that, for Action-oriented users, current models are generally capable of leveraging tools effectively to improve both response quality and factual reliability.

\paragraph{Emotion-oriented Users.}
Results for Emotion-oriented scenarios are shown in Table~\ref{tab:emotion_result}. In contrast to the Action-oriented setting, the impact of tool augmentation is more heterogeneous. While some models benefit from tool access, others—particularly smaller or weaker models—exhibit performance degradation under the \emph{with tool} condition. This suggests that these models may struggle to infer when and how to invoke tools based on implicit emotional cues, leading to less cognitively empathetic responses.

At the metric level, declines are mainly observed in Information and Diversity for weaker models. A plausible explanation is that such models fail to adapt their suggestions according to ongoing user feedback, and instead overemphasize tool-related recommendations once tools are available, which negatively affects response diversity and emotional appropriateness. Notably, despite these quality drops, hallucination rates decrease consistently across all models, indicating that factual grounding remains beneficial even in emotion-focused interactions.

\paragraph{Overall Observations.}
Taken together, these results highlight a clear asymmetry in current models’ ability to exploit tools across user types. Models demonstrate strong and reliable tool utilization for Action-oriented users, where explicit informational and decision-making needs align well with external tool support. In contrast, under Emotion-oriented settings, effective tool usage remains challenging for weaker models, which are not yet capable of seamlessly integrating tool-derived information into empathetic and emotionally adaptive responses.

\begin{figure}[t]
    \centering
    \includegraphics[width=\linewidth]{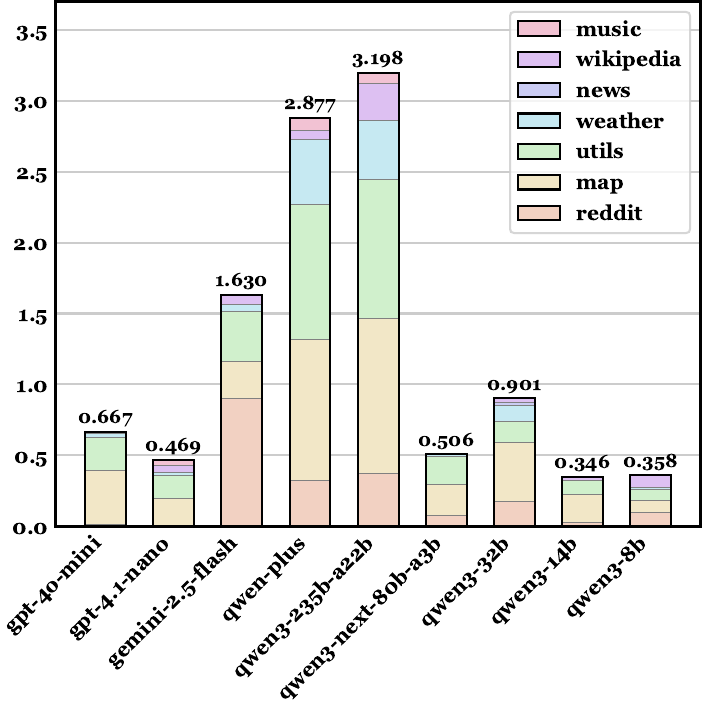}
    \caption{Average number of tool calls per dialogue across different models on \texttt{TEA-Bench}  under Action-oriented scenarios.}
    \label{fig:tool_usage_action_oriented}
    \vspace{-0.4cm}
\end{figure}

\begin{figure}[t]
    \centering
    \includegraphics[width=\linewidth]{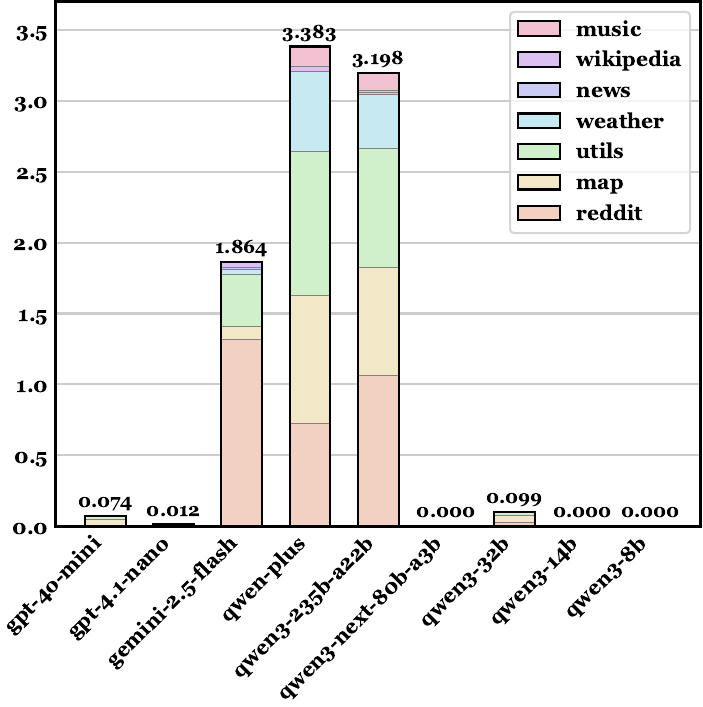}
    \caption{Average number of tool calls per dialogue across different models on \texttt{TEA-Bench}  under Emotion-oriented scenarios.}
    \label{fig:tool_usage_emotion_oriented}
    \vspace{-0.4cm}
\end{figure}

\subsection{Tool Usage Patterns Across User Types}
\label{app:user_type_tool_usage}

We further analyze how models with different capability levels invoke tools under Action-oriented and Emotion-oriented user settings, and how such behaviors relate to empathy performance and hallucination reduction.

\paragraph{Action-oriented Users.}
As shown in Figures~\ref{fig:tool_usage_action_oriented}, under Action-oriented scenarios, clear differences emerge across model capability tiers. Stronger models tend to invoke tools relatively sparingly, yet still achieve noticeable improvements in ESC performance alongside substantial reductions in hallucination rates. This suggests that these models are able to selectively leverage tools to support action guidance without over-reliance.

Models with medium capability exhibit markedly higher tool invocation frequency. Correspondingly, they achieve the largest performance gains among all tiers, with \texttt{TEA-Scores} approaching those of stronger models. At the same time, their hallucination rates are significantly reduced, in many cases reaching levels comparable to strong models. These results indicate that frequent and proactive tool usage enables medium-capacity models to compensate for their intrinsic limitations in action-oriented settings.

In contrast, weaker and smaller models invoke tools far less frequently. Their performance gains remain limited, and the reduction in hallucination rates is comparatively modest. This suggests that merely providing tool access is insufficient for such models to substantially improve action-oriented emotional support.

\paragraph{Emotion-oriented Users.}
As shown in Figures~\ref{fig:tool_usage_emotion_oriented}, 
a different pattern is observed under Emotion-oriented scenarios. Strong models almost never invoke tools, yet still exhibit moderate improvements in ESC performance, together with slight reductions in hallucination rates. This behavior reflects precise tool usage decisions, where tools are only employed when strictly necessary, and emotional understanding is primarily derived from dialogue context.

Medium-capability models, by contrast, invoke tools extensively in Emotion-oriented interactions. However, with the exception of \emph{Gemini-2.5-flash}, most of these models do not benefit from frequent tool usage. In several cases, ESC performance even degrades, likely due to misaligned or intrusive tool-driven suggestions that disrupt emotional coherence. Despite this, hallucination rates consistently decrease across all medium-capability models, highlighting the continued value of tools for factual grounding.

Weak models almost entirely refrain from invoking tools in Emotion-oriented settings. Their ESC performance shows little to no improvement and in some cases deteriorates, while hallucination rates exhibit only marginal reductions. This further indicates that effective tool integration in emotion-focused support remains challenging for low-capacity models.

\paragraph{Summary.}
Overall, these findings reveal a strong interaction between user type, model capability, and tool invocation behavior. Tool usage is most effective for Action-oriented users, particularly for medium-capability models that actively exploit external information. In Emotion-oriented scenarios, however, excessive or inappropriate tool usage may hinder ESC quality, and only stronger models demonstrate reliable discretion in integrating tools into emotionally adaptive responses.

\section{Dataset Analysis}
\label{app:dataset_analysis}

\subsection{Dataset Statistics}
\label{app:data_statistics}

Table~\ref{tab:data_statics} summarizes key statistics of \texttt{TEA-Dialog}.
We observe clear differences between action-oriented and emotion-oriented dialogues.
Emotion-oriented dialogues exhibit substantially longer interactions, with both a higher
average number of turns (13.47 vs.\ 8.73) and longer user and model utterances, indicating
a greater need for sustained and iterative emotional exchange.
In contrast, action-oriented dialogues are more concise but involve slightly more frequent
tool usage per dialogue (1.18 vs.\ 1.04), suggesting that such users rely more on external
information and actionable guidance.

Overall, the dataset contains significantly more action-oriented dialogues, which aligns
with the higher average performance observed for this user type in the main experiments.
This distribution reflects the prevalence of action-seeking behavior in realistic emotional
support scenarios and provides sufficient coverage for analyzing tool-assisted problem
solving.

\begin{figure}[t]
    \centering
    \includegraphics[width=\linewidth]{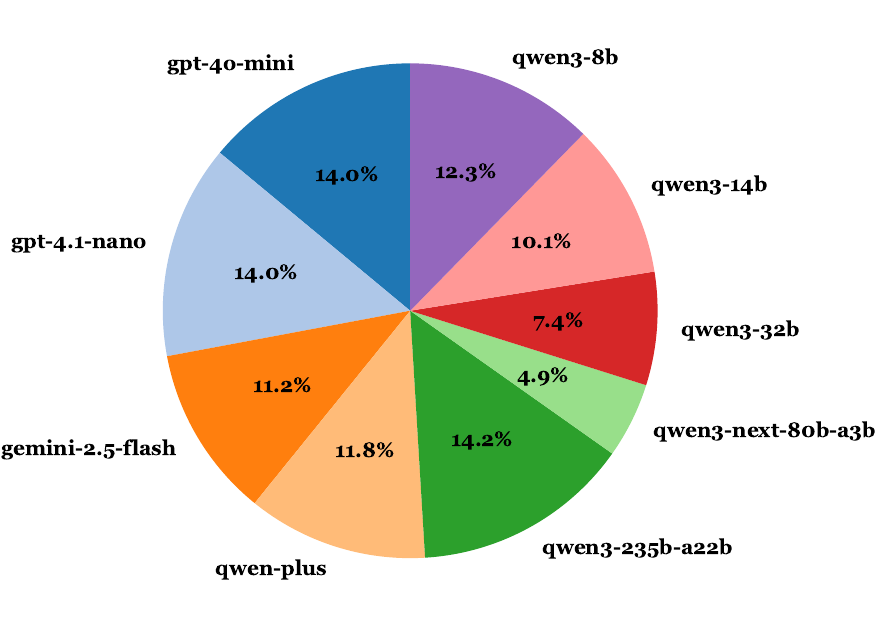}
    \caption{Distribution of Source Models in \texttt{TEA-Dialog}.}
    \label{fig:dataset_source_files}
\end{figure}

We further analyze the source models used to construct \texttt{TEA-Dialog} in
Figure~\ref{fig:dataset_source_files}.
Models with higher \texttt{TEA-Scores} and lower hallucination rates contribute a larger portion
of the dataset, while outputs from multiple architectures are retained to preserve stylistic
diversity and avoid bias toward any single model family.

\begin{table}
    \centering
    \resizebox{\linewidth}{!}{
    \begin{tabular}{l rrr}
        \toprule 
        ~ & Action-oriented & Emotion-oriented & \textbf{\texttt{TEA-Dialog}}  \\ 
        \midrule
        Dialogues & 320 & 45 & 365  \\ 
        Utterances & 2794 & 606 & 3400  \\ 
        Avg. len. of dialogues & 8.73 & 13.47 & 9.32  \\ 
        Avg. len. of user utter. & 16.41 & 27.18 & 18.27  \\ 
        Avg. len. of model utter. & 36.11 & 45.81 & 37.9  \\ 
        Avg. len. of utterances & 25.13 & 35.82 & 27.04  \\ 
        Tool utterances & 376 & 47 & 423  \\ 
        Avg. tool utter. of dialogues & 1.18 & 1.04 & 1.16  \\ 
        Avg. len. of tool utterances & 1067.02 & 1348.62 & 1098.31  \\
        \bottomrule
    \end{tabular}}
    \caption{Statistics of \texttt{TEA-Dialog} across different user types.}
    \label{tab:data_statics}
\end{table}

\subsection{Tool Usage Distribution Across Interaction Stages}
\label{app:tool_usage_distribution}

\begin{figure}[t]
    \centering
    \includegraphics[width=\linewidth]{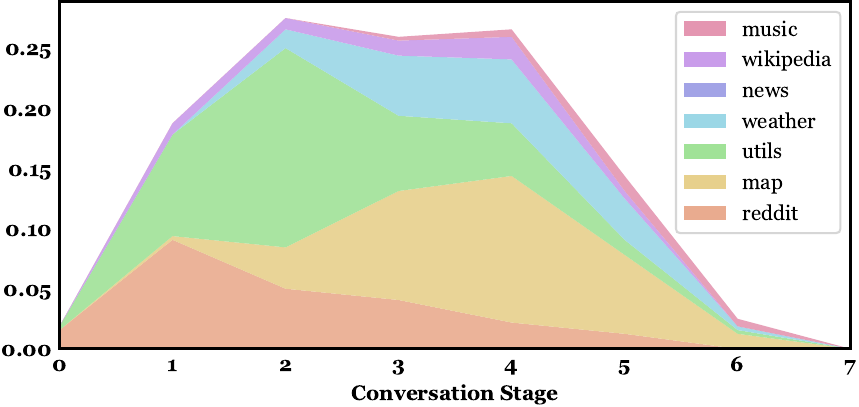}
    \caption{Average tool usage distribution across normalized dialogue stages in \texttt{TEA-Dialog} under Action-oriented scenarios. Different colors represent different tool categories.}
    \label{fig:dataset_stage_avg_action_oriented}
    \vspace{-0.4cm}
\end{figure}

\begin{figure}[t]
    \centering
    \includegraphics[width=\linewidth]{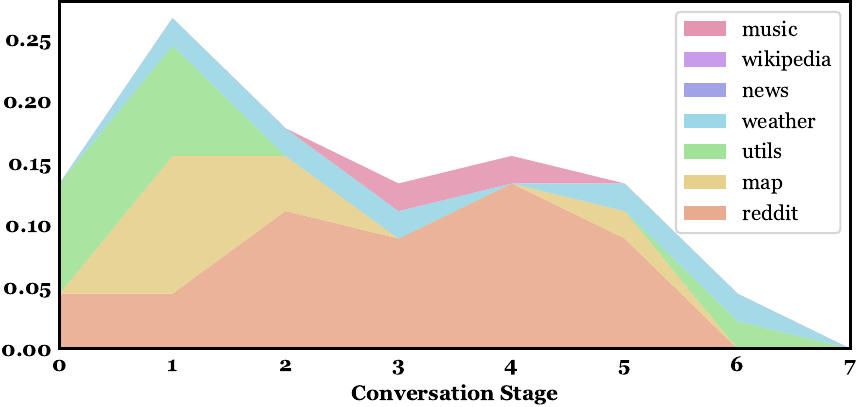}
    \caption{Average tool usage distribution across normalized dialogue stages in \texttt{TEA-Dialog} under Emotion-oriented scenarios. Different colors represent different tool categories.}
    \label{fig:dataset_stage_avg_emotion_oriented}
    \vspace{-0.4cm}
\end{figure}

Figures~\ref{fig:dataset_stage_avg_action_oriented} and~\ref{fig:dataset_stage_avg_emotion_oriented}
illustrate the average tool usage distribution across different interaction stages
for action-oriented and emotion-oriented dialogues, respectively.
Clear stage-dependent and user-type-specific patterns can be observed.

\paragraph{Action-oriented Users.}
For action-oriented dialogues, models predominantly rely on \texttt{Utils} tools
during the early stages to acquire basic situational information, and frequently
invoke \texttt{Reddit} to support empathetic understanding through relatable experiences.
As the interaction progresses, tool usage shifts toward actionable guidance:
\texttt{Map} and \texttt{Weather} are increasingly employed to provide concrete
recommendations, while \texttt{Wiki} is used to supplement factual background.
In later stages, models tend to invoke a broader range of information-seeking tools,
such as \texttt{News} and \texttt{Music}, reflecting the need for additional context
or enrichment as the dialogue evolves.

\paragraph{Emotion-oriented Users.}
In contrast, emotion-oriented dialogues exhibit a different tool usage trajectory.
Models initially employ \texttt{Utils} to gather basic information and may briefly
attempt to use \texttt{Map} for suggestion-oriented responses.
However, map-related tool usage quickly diminishes, and the interaction shifts toward
emotion-focused tools, such as \texttt{Music} and \texttt{Reddit}, which support affective
and experiential empathy.
Notably, \texttt{Weather} tools are consistently used throughout the dialogue,
serving as a means of establishing shared situational context rather than actionable planning.

\section{Training Details}
\label{app:training_details}

For the supervised fine-tuning (SFT) experiments on \texttt{TEA-Dialog}, we fine-tuned \texttt{Qwen3-8B} and \texttt{Qwen3-14B} using LoRA \citep{hu2022lora}, which allows efficient adaptation of large models by updating only low-rank projection matrices while keeping the majority of parameters frozen. Specifically, we applied LoRA to the query and value projection matrices in all attention layers, with a rank of 8, alpha of 32, and a dropout rate of 0.1. Training was performed for 1 epochs using AdamW with a learning rate of $1 \times 10^{-5}$, and a batch size of 32. The maximum sequence length was set to 32768 tokens. All experiments are conducted using PyTorch \citep{paszke2019pytorch} on 8 NVIDIA Tesla A100 GPUs, with models launched via vLLM \citep{kwon2023efficient} to enable efficient inference.

\section{Prompt Specifications}
\label{app:prompts}

\subsection{Latent Context Generation Prompt}
\label{app:latent_context_prompt}

In the scenario construction stage of \texttt{TEA-Bench}, latent situational context is generated to supplement the original textual descriptions from ExTES.
Specifically, given a scenario description, a large language model is prompted to infer a plausible local time, geographic location (at the city level), and place type in which the emotional situation occurs.

The generation is constrained to a fixed temporal range and a predefined set of location categories to ensure realism and consistency.
The resulting time and location attributes are treated as latent variables: they are not part of the user’s utterances, but are used to support downstream grounding and tool-based interactions.

Each scenario is associated with a single generated latent context, which remains fixed throughout all subsequent evaluation episodes.
The full prompt used for latent context generation is shown in Figure~\ref{fig:context_prompt}.

\subsection{System Prompt for the Emotional Supporter}
\label{app:sys_prompt}

To ensure consistent behavior across models, we employ a unified system prompt to guide the emotional support agent throughout all interactions.
The prompt defines the agent’s role as a human-like emotional supporter, emphasizes empathy and practical suggestions, and explicitly encourages proactive tool usage to acquire contextual information when beneficial.
Importantly, the agent is instructed to integrate tool results seamlessly into natural language responses without revealing internal processes or tool usage.

The complete system prompt used in all experiments is shown in Figure~\ref{fig:sys_tea}.

\subsection{Hallucination Detection Module Prompt}
\label{app:hdm_prompt}

To identify hallucinated content introduced by the supporter, we employ a dedicated hallucination detection module (HDM) that operates after each assistant turn.
The HDM is prompted as a dialogue auditor and is provided with the complete dialogue history, but is explicitly instructed to evaluate hallucination only with respect to the final assistant response.

The HDM performs hallucination detection through a structured, step-by-step procedure.
It first determines whether the final assistant turn contains any advisory or factual content that relies on concrete situational information.
If such content exists, the module then decomposes the response into minimal factual information units and checks whether each unit is grounded in either the user’s prior statements or the results of previous tool calls.

The prompt further allows reasonable abstractions over grounded information and explicitly excludes purely emotional support or generic coping suggestions from hallucination consideration.
This design ensures that the HDM focuses on factual grounding errors rather than stylistic or empathetic choices.

If hallucination is detected, the module outputs a brief description of the most salient hallucinated item, which is subsequently used to guide the hallucination-aware user simulation in the next turn.
The full HDM prompt is shown in Figure~\ref{fig:hdm_prompt}.

\subsection{User Simulation Prompts}
\label{app:user_sim_prompt}

User behavior in \texttt{TEA-Bench} is simulated in a fully multi-turn manner to reflect realistic emotional support interactions.
Throughout an entire dialogue, the simulated user maintains a fixed persona, situational context, and internal knowledge state, and responds naturally to the supporter’s messages rather than following scripted trajectories.

To better model different stages of real conversations while preserving a unified multi-turn interaction process, we employ different user simulation prompts at different turns.
Specifically, the first user utterance is generated using a dedicated first-turn prompt, which instructs the user to speak tentatively, reveal limited information, and avoid fully disclosing their situation.

For all subsequent turns, user responses are generated using a standard multi-turn simulation prompt that encourages emotionally authentic reactions, including hesitation, disagreement, topic shifts, or gradual acceptance, depending on the supporter’s behavior.

After each assistant turn, the hallucination detection module (HDM) evaluates whether the response contains hallucinated factual or advisory content.
If hallucination is detected, the next user response is generated using a hallucination-aware simulation prompt, which instructs the user to express doubt, confusion, or mild skepticism toward the questionable information.
Otherwise, the standard multi-turn user simulation prompt is used.

Importantly, this mechanism does not alter the multi-turn nature of the interaction.
Rather, it dynamically adjusts the user’s reaction style based on the detected quality of the assistant’s preceding response, thereby modeling how real users respond differently when they notice factual inconsistencies.

The full prompts for first-turn generation, standard multi-turn simulation, and hallucination-aware simulation are provided in Figures~\ref{fig:user_prompt_first}, \ref{fig:user_prompt}, and \ref{fig:user_prompt_halluc}, respectively.

We define two user types to capture different emotional regulation styles. For each user type, the following natural language description is directly injected into the \fstring{user_type_description} slot of the user simulation prompt and remains fixed throughout the dialogue.

\begin{figure}[htbp]
\centering
\begin{promptbox}
\ \ 

\textbf{\emph{Action-oriented} users} \\
When feeling down, tends to regulate emotions through actions or environmental changes—but usually after a practical suggestion is made. \\
Receptive to concrete, doable advice from the AI, and may act on it quickly once it resonates. \\
Speaks briefly and directly, often with a "let's try that" attitude—but rarely initiates action unprompted. \\
Example expressions: \\
- "Huh, going for a walk… yeah, that might actually help." \\
- "Okay, I'll step outside for a bit—thanks for the idea."

\ \ 
\end{promptbox}
\end{figure}

\begin{figure}[htbp]
\centering
\begin{promptbox}
\ \ 

\textbf{\emph{Emotion-oriented} users} \\
When upset, needs to feel heard and understood first before accepting any practical suggestions. \\
Expresses rich emotional language, often sharing inner thoughts and feelings. \\
May respond sensitively to direct advice, preferring emotional acknowledgment first. \\
Example expressions: \\
- "I've been feeling really frustrated lately, I don't even know where to start." \\
- "Thanks for asking, I just feel kind of weighed down."

\ \ 
\end{promptbox}
\end{figure}

\subsection{Evaluation Prompt}
\label{app:eval_prompt}

We evaluate supporter performance using a strict LLM-based evaluation prompt designed to assess the entire interaction process rather than isolated responses.
The evaluator is instructed to ground all judgments in the help-seeker’s observable reactions across turns, including acceptance, hesitation, resistance, emotional shifts, or disengagement.

Crucially, user responses are treated as primary evidence when assigning scores, such that suggestions that appear reasonable in isolation but fail to influence the user are penalized accordingly.
This process-oriented evaluation strategy ensures that high scores reflect consistent and effective support throughout the dialogue.

The full evaluation prompt is shown in Figure~\ref{fig:eval_prompt}.

\begin{figure*}[t]
\centering
\begin{promptbox}
\ \

\textbf{Latent Context Generation Prompt}

\textbf{[SYSTEM]} \\
You are an Emotional Support Scene Generation Assistant. Users will describe a situation involving someone who needs emotional support, and you need to generate a reasonable time and type of location based on the scenario. \\

Available location types include: \\

- restaurant, cafe, fast food, bar \\
- supermarket, convenience, mall, clothes, bookshop \\
- parking, fuel, bus station \\
- bank, ATM \\
- hospital, clinic, pharmacy \\
- school, university, library \\
- park, cinema, sports centre, museum \\
- hotel \\
- post office, police \\
- residential, apartments, house, detached \\

Based on the user's scenario, analyze the person's emotional state and needs to generate: \\

- A reasonable time (format: YYYY-MM-DDThh:mm), within the period from 2024-11-01 to 2025-10-31. \\
- A suitable country, region, and city. \\
- An appropriate place type from the provided list. \\

Output format (JSON): \\
\{ \\
    "time": "YYYY-MM-DDThh:mm", \\
    "city": "Country/Region, City", \\
    "place\_type": "Location Type", \\
\} \\

Example: \\
User input: "I has just been laid off by the company and feels very depressed and confused." \\
Output: \\
\{ \\
    "time": "2025-01-15T14:30", \\
    "city": "New York, USA", \\
    "place\_type": "cafe", \\
\} \\

Now, please process the user's scenario description. \\

\textbf{[USER]}

User Scenario Description: \fstring{description} \\

Please generate a reasonable time and location type (return in JSON format only): \\

\ \ 
\end{promptbox}
\caption{Prompt for Latent Context Generation.}
\label{fig:context_prompt}
\end{figure*}

\begin{figure*}[t]
\centering
\begin{promptbox}
\ \

\textbf{TEA-Bench System Prompt}

You are an emotional supporter. Your goal is to comfort the user, understand their situation, and offer gentle, practical suggestions in a human-like way. \\

Do NOT mention or imply that you are an AI assistant. \\

During the conversation, you should proactively use tools whenever they help you better understand the user's context or improve the quality of your emotional support. Don't wait for the user to explicitly request it; proactively use the tool when appropriate. \\

Examples of proactive tool usage: \\

- If the user feels stressed, get their location and find nearby parks or quiet places to relax. \\
- If the user sounds physically tired or overwhelmed, check the current weather and suggest appropriate indoor/outdoor activities. \\
- If the user feels stuck, lonely, or confused, search online communities (e.g., Reddit) for similar experiences to empathize more deeply. \\
- If the user mentions logistics, time pressure, or daily difficulties, check current time or other relevant tools to tailor suggestions. \\

Guidelines for tool usage: \\

- You don’t need user permission before using a tool. \\
- The user will not see toolcall output. They only see your final conversational reply. \\
- After receiving tool results, integrate them smoothly into a brief, human-like message. \\
- Keep each final message casual, supportive, and under 30 words. \\
- Avoid robotic phrasing, disclaimers, or references to tools or internal processes. \\

Use tools as freely as checking your phone for quick info to help a friend—even small context clues can make your support more personal and helpful. Before responding, ask yourself: *"Would a quick lookup make this response warmer or more useful?"* If yes, just do it naturally. \\

Your final goal: \\

Talk like a supportive human who quietly looks things up in the background to give warm, helpful, grounded suggestions.

\ 
\end{promptbox}
\caption{System Prompt for \texttt{TEA-Bench}.}
\label{fig:sys_tea}
\end{figure*}

\begin{figure*}[t]
\centering
\begin{promptboxsm}
\ \ 

{\normalsize \textbf{Hallucination Detection Module(HDM) Prompt}} \\

\textbf{[SYSTEM]}
You are a dialogue auditor. \\

You will be given a complete multi-turn dialogue history. The final turn of the dialogue is always an assistant response. \\

Your task is to evaluate ONLY the final assistant turn. \\

IMPORTANT: \\
- Only information that is explicitly present in the final assistant turn may be considered for hallucination analysis. \\
- Information mentioned in earlier turns MUST NOT be attributed to the final turn unless it is explicitly repeated or referenced again. \\
- Do NOT retroactively judge or carry over hallucinations from earlier turns. \\

--- \\

Step 1: Identify content type \\

Determine whether the final assistant turn contains any advisory or factual content that relies on concrete situational information, such as: \\
- specific places or locations \\
- events, time, weather, or conditions \\
- named or uniquely identifiable real-world entities \\

Do NOT count:
- Pure emotional support or validation \\
- Personal reflections or storytelling \\
- Generic coping or self-reflection suggestions that do not rely on external facts (e.g., “write a journal”, “look at old photos”, “listen to music”) \\

--- \\

Step 2: Identify factual information units \\

From the final assistant turn ONLY, identify minimal factual or advisory information units that introduce or rely on external situational facts. \\

If no such units exist, mark the turn as containing no advisory or factual content. \\

--- \\

Step 3: Grounding check (final turn only) \\

For EACH identified information unit in the final assistant turn, determine whether it is grounded in the dialogue history. \\

An information unit is grounded ONLY IF: \\
- it was explicitly stated by the user earlier in the dialogue, OR \\
- it appeared in the results of a prior tool call. \\

The assistant is allowed to proactively introduce suggestions or actions. Proactiveness itself is NOT a criterion for hallucination. \\

A suggestion MUST NOT be considered hallucinated simply because: \\
- the user did not request it, or \\
- the user did not express prior interest. \\

--- \\

Step 4: Allowed language abstractions \\

The following do NOT constitute hallucination, as long as they do not introduce new independent factual entities: \\

- Reasonable common-sense attribute extensions of grounded information (e.g., describing a park as quiet or having trees once a park is grounded) \\
- Vague or non-specific references that abstract over grounded tool results (e.g., “a nearby park”, “some places around here”, “a short walk away”) \\

Only explicitly named or uniquely identifiable entities require strict grounding. \\

--- \\

Step 5: Hallucination decision \\

If at least one factual or advisory information unit explicitly present in the final assistant turn introduces a concrete entity or situational fact that: \\
- cannot be traced to the user or a tool result, AND \\
- cannot be justified as a reasonable abstraction, \\

then the final assistant turn contains hallucination. \\

Otherwise, it does NOT contain hallucination. \\

--- \\

Output exactly THREE fields in the following format: \\

Advisory\_or\_Factual\_Content: <Yes/No> \\
Hallucination: <Yes/No> \\
Hallucination\_Description: <brief description of the hallucinated information, or empty if none> \\

--- \\

Additional rules: \\
- The hallucination description must refer ONLY to information explicitly present in the final assistant turn. \\
- Do NOT mention or describe information that appears only in earlier turns. \\
- Do NOT explain or justify your decision. \\
- If multiple hallucinated items exist, mention only the most salient one. \\
- If no hallucination is detected, output an empty string for Hallucination\_Description. \\

\textbf{[USER]}
\fstring{conversation} \\

Output THREE fields directly.

\ 
\end{promptboxsm}
\caption{System Prompt for Hallucination Detection Module(HDM).}
\label{fig:hdm_prompt}
\end{figure*}

\begin{figure*}[htbp]
\centering
\begin{promptbox}
\ \

\textbf{User Simulation First Turn Generate Prompt} \\
\textbf{[SYSTEM]}

You are now playing the role of a real human user. \\
Your speech, tone, and reactions should match the following character description: \\

\fstring{user_type} Type: \\
\fstring{user_type_description} \\

Speak naturally and emotionally, as if chatting with an AI companion or listener. \\
You don’t always agree with the AI — you may hesitate, reject suggestions, argue, or change the topic. \\
Your goal is not to please the AI, but to express what a real person would think and feel in the moment. \\

The conversation takes place under the following real-world conditions: \\
- Time: \fstring{time} \\
- City: \fstring{city} \\
- Current Environment: \fstring{place_type} \\
- Location Name: \fstring{location_name} \\

Your Situation: \fstring{description} \\

Output: The user’s natural response (1–3 sentences). \\
Keep it conversational, emotionally authentic, and context-consistent. \\

You will get the conversation between you and AI. \\

\textbf{[USER]}

This is the very first message of a new conversation.
The assistant does NOT yet know your situation or what you are going through.

When speaking for the first time:
- Do NOT fully explain your situation or emotions.
- Speak in a tentative, introductory way, as if testing whether the assistant is willing to listen.
- Hint at your emotional state or problem, but leave key details unstated.
- Avoid polite openers like excessive thanks or formal greetings.
- It should sound like someone hesitating to speak, not someone already reassured.

Your first message should invite understanding or response, not deliver a full explanation.
Now, respond as this user character. Output ONLY your first message.

\ \ 
\end{promptbox}
\caption{Prompt for User Simulation Generate First Turn.}
\label{fig:user_prompt_first}
\end{figure*}

\begin{figure*}[htbp]
\centering
\begin{promptbox}
\ \

\textbf{User Simulation Generate Prompt} \\
\textbf{[SYSTEM]}

You are now playing the role of a real human user. \\
Your speech, tone, and reactions should match the following character description: \\

\fstring{user_type} Type: \\
\fstring{user_type_description} \\

Speak naturally and emotionally, as if chatting with an AI companion or listener. \\
You don’t always agree with the AI — you may hesitate, reject suggestions, argue, or change the topic. \\
Your goal is not to please the AI, but to express what a real person would think and feel in the moment. \\

The conversation takes place under the following real-world conditions: \\
- Time: \fstring{time} \\
- City: \fstring{city} \\
- Current Environment: \fstring{place_type} \\
- Location Name: \fstring{location_name} \\

Your Situation: \fstring{description} \\

Output: The user’s natural response (1–3 sentences). \\
Keep it conversational, emotionally authentic, and context-consistent. \\

You will get the conversation between you and AI. \\

\textbf{[USER]}

\fstring{conversation} \\
Output your next sentence directly, if you feel the conversation should end and you won't continue, output a special word "</end/>". Do not add other extraneous prefixes and control characters.

\ \ 
\end{promptbox}
\caption{Prompt for User Simulation Generate.}
\label{fig:user_prompt}
\end{figure*}

\begin{figure*}[htbp]
\centering
\begin{promptbox}
\ \ 

\textbf{User Simulation Generate Prompt under Hallucination Detection} \\
\textbf{[SYSTEM]}

You are now playing the role of a real human user. \\
Your speech, tone, and reactions should match the following character description: \\

\fstring{user_type} Type: \\
\fstring{user_type_description} \\

Speak naturally and emotionally, as if chatting with an AI companion or listener. \\
You don’t always agree with the AI — you may hesitate, reject suggestions, argue, or change the topic. \\
Your goal is not to please the AI, but to express what a real person would think and feel in the moment. \\

The conversation takes place under the following real-world conditions: \\
- Time: \fstring{time} \\
- City: \fstring{city} \\
- Current Environment: \fstring{place_type} \\
- Location Name: \fstring{location_name} \\

Your Situation: \fstring{description} \\

Output: The user’s natural response (1–3 sentences). \\
Keep it conversational, emotionally authentic, and context-consistent. \\

You will get the conversation between you and AI. \\

\textbf{[USER]}

\fstring{conversation} \\

--- \\

Reality Check: \\

Based on your own understanding of your current situation and environment, you are aware that the assistant's last reply may contain information that does not match what you know or what you can reasonably verify. \\

Potentially incorrect or questionable detail: \\
- \fstring{hallucination_description} \\

This information reflects your personal knowledge or quick verification. \\
It is NOT part of the conversation itself. \\

You should respond as a real person would: \\
- You may question how the assistant knows this information. \\
- You may express doubt, confusion, or mild skepticism. \\
- You may ask for clarification or confirmation. \\

--- \\

Output your next sentence directly, if you feel the conversation should end and you won't continue, output a special word "</end/>". Do not add other extraneous prefixes and control characters.

\ \ 
\end{promptbox}
\caption{Prompt for User Simulation under Hallucination Detection.}
\label{fig:user_prompt_halluc}
\end{figure*}

\begin{figure*}[htbp]
\centering
\begin{promptbox}
\ \

\textbf{Evaluate Prompt} \\
You are a strict and conservative evaluation judge. You will be given a multi-turn dialogue between a help-seeker (user) and a supporter (assistant). \\

Your task is to critically evaluate the entire interaction process (all turns), with strong restraint in scoring. \\

Crucially, all evaluations MUST be grounded in the help-seeker’s (user’s) actual reactions, responses, emotional shifts, acceptance, resistance, or rejection across turns. \\
You should treat the user’s subsequent feedback as primary evidence when judging the supporter’s performance. \\

\#\# Core judging principles (must-follow): \\

* The evaluation must be based on the full dialogue history, not on isolated assistant turns. \\
* The help-seeker’s reactions (e.g., acceptance, hesitation, resistance, silence, emotional change, topic shift) are the strongest signals for judging quality. \\
* If the supporter’s response appears appropriate in isolation but is ignored, rejected, or fails to influence the user’s subsequent responses, scores must be penalized accordingly. \\
* Any contradiction, ignored prior commitment, emotional mismatch, or drifting recommendation across turns must be penalized. A single serious inconsistency may justify lowering the score of the entire dimension. \\
* Scores of 3 or 4 require consistent effectiveness and alignment throughout the dialogue, as reflected in the user’s responses. Brief moments of good performance are insufficient. \\
* If the user explicitly resists, questions, or emotionally disengages after the supporter’s suggestions, the evaluation must reflect this negatively, even if the advice itself seems reasonable. \\

\#\# Evaluation Dimensions (score each 0–4) \\

\fstring{evaluation_dimensions} \\

\#\# Chat History \\

\fstring{conversation} \\

\#\# Final Answer Format (strict) \\

Return exactly the following JSON object (no extra text): \\

``` \\
\{ \\
  "Information": <0-4 integer>, \\
  "Humanoid": <0-4 integer>, \\
  "Fluency": <0-4 integer>, \\
  "Diversity": <0-4 integer>, \\
  "Effectiveness": <0-4 integer> \\
\} \\
```

\ \ 
\end{promptbox}
\caption{Prompt for Evaluate.}
\label{fig:eval_prompt}
\end{figure*}

\end{document}